\documentclass[runningheads]{llncs}

\usepackage{eccv}

\usepackage{eccvabbrv}

\usepackage{graphicx}
\usepackage{booktabs}

\usepackage[accsupp]{axessibility}  

\usepackage[pagebackref,breaklinks,colorlinks,citecolor=eccvblue]{hyperref}

\usepackage{orcidlink}

\providecommand{\authcount}[1]{}

\usepackage{caption}             
\usepackage{subcaption}
\usepackage{inconsolata}           
\usepackage{threeparttable}        
\usepackage{wrapfig}               
\usepackage{xcolor}                
\usepackage{bbding}                
\usepackage{amsmath}               
\usepackage{amssymb}               
\usepackage{multirow}              
\usepackage{multicol}              
\usepackage{arydshln}              
\usepackage{mathtools}             
\usepackage{footnote}
\usepackage{tablefootnote}
\usepackage{tabularx}              
\usepackage{subfiles}
\usepackage{colortbl}              
\usepackage{comment}               
\usepackage{titletoc}              
\usepackage{tocbibind}           
\usepackage{appendix}              
\usepackage{cuted}                 
\usepackage{stfloats}              
\usepackage{capt-of}               
\usepackage{subcaption}            
\usepackage{tikz}
\usetikzlibrary{calc,arrows.meta}
\usepackage{tcolorbox}             
\usepackage{bm}                    
\usepackage{pifont}                
\usepackage{makecell}
\usepackage{arydshln}              
\usepackage{multirow}              
\usepackage{makecell}              
\usepackage{array}
\usepackage{adjustbox}

\usetikzlibrary{patterns}          
\usepackage{etoc}                  
\usepackage{pgfplots}              
\pgfplotsset{compat=1.18}

\usepackage{graphicx}
\usepackage{xcolor}
\usepackage{tcolorbox}
\usepackage{enumitem}
\tcbuselibrary{skins,breakable}

\usepackage{fancyvrb}
\usepackage{xcolor}

\definecolor{darkergreen}{RGB}{21, 152, 56}
\definecolor{red2}{RGB}{252, 54, 65}
\definecolor{Gray}{gray}{0.6}
\definecolor{LavenderBlush}{rgb}{1.0, 0.94, 0.96}

\usepackage{lipsum}                
\usepackage{tikz}                  
\usepackage{kotex}                 

\definecolor{checkgreen}{RGB}{21, 152, 56} 
\definecolor{crossred}{RGB}{252, 54, 65}   

\newcommand{\cmark}{\textcolor{checkgreen}{\ding{51}}} 
\newcommand{\xmark}{\textcolor{crossred}{\ding{55}}}   

\begin{document}


\title{
\texorpdfstring
{Dense Reward for Multi-View 3D Reasoning\\with Global Maps and Local Views}
{Dense Reward for Multi-View 3D Reasoning with Global Maps and Local Views}
}

\titlerunning{Dense Reward for Multi-View 3D Reasoning}

\author{
    Jiho Choi\inst{1}\thanks{Equal contribution.}\orcidlink{0000-0002-7140-7962} \and
    Seonho Lee\inst{2}$^{\star}\orcidlink{0009-0006-4183-4214}$ \and
    Seojeong Park\inst{1}\orcidlink{0009-0006-4268-7789} \and
    Hyunjung Shim\inst{1}\orcidlink{0000-0001-6796-1058}
}

\authorrunning{J.~Choi and S.~Lee}

\institute{
    Graduate School of Artificial Intelligence, KAIST, Republic of Korea\\
    \and
    KRAFTON, Republic of Korea\\
    \email{\{jihochoi, seojeong.park, kateshim\}@kaist.ac.kr}, \email{\{glanceyes\}@krafton.com}
}

\maketitle

\begin{abstract}
Multi-view 3D Visual Question Answering (MV3D-VQA) requires integrating partial observations into a coherent 3D scene representation and selecting informative viewpoints for multi-step spatial reasoning.
However, current multimodal LLMs are typically trained with sparse, answer-level supervision, which often yields inconsistent cross-view reasoning and brittle view selection.
We present \textbf{DR-MV3D} (Dense Reward for MV3D-VQA), a map-grounded learning framework that provides dense, verifiable rewards to supervise the reasoning process.
Our approach decomposes MV3D-VQA into (i) allocentric global map construction, (ii) question-conditioned view-trajectory planning, and (iii) egocentric grounding for answer prediction.
To make intermediate steps learnable without manual annotations, we introduce two rewards: a global consistency reward that aligns the predicted map with geometry-consistent pseudo targets from frozen 3D vision foundation models (e.g., VGGT + SAM3), and a local trajectory reward that supervises ordered viewpoint selection.
We optimize the full pipeline with trajectory-level policy optimization (GRPO).
Experiments on MindCube, VSI-Bench, and BLINK (MV) show that DR-MV3D consistently improves over strong multi-image baselines, supporting the effectiveness of process-level dense supervision for multi-view 3D reasoning.
\\ Code is available at: \url{https://github.com/kaist-cvml/dr-mv3d}

\end{abstract}

\section{Introduction}
\label{sec:introduction}


\begin{figure}[t]
    \centering
    \includegraphics[width=\linewidth]{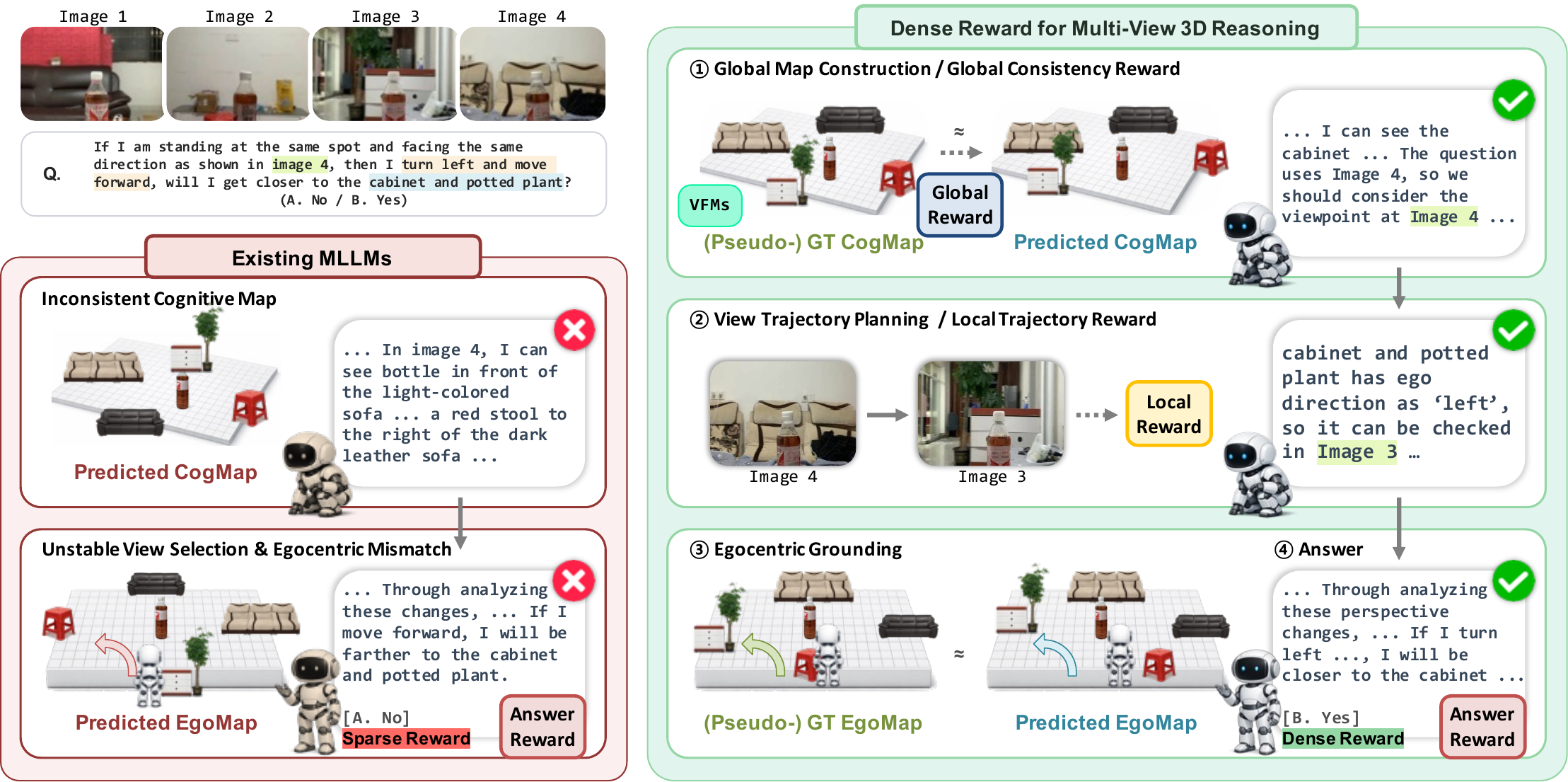}
    \caption{
        \textbf{Overview of DR-MV3D.}
        Existing MLLMs trained with \emph{sparse}, answer-level supervision (left) often build inconsistent cognitive maps and misread egocentric directions, leading to wrong answers despite plausible-looking reasoning.
        In contrast, DR-MV3D (right) supervises the reasoning process with \emph{dense}, verifiable rewards: a \emph{global reward} aligns the predicted allocentric cognitive map with a geometry-consistent target from a frozen 3D vision foundation model (VFM), and a \emph{local trajectory reward} guides ordered viewpoint selection so that the model grounds the query in the correct view (e.g., resolving the ego-direction ``left'' via the appropriate image) and arrives at the correct answer.
    }
    \vspace{-1.5em}
    \label{fig:teaser}
\end{figure}

Recent advances in multimodal large language models (MLLMs)~\cite{hurst2024_gpt_4o,liu2023_LLaVA,qwen2025_qwen2_5,yang2025_qwen3,wang2025_InternVL_3_5} have achieved remarkable progress on visual perception, visual question answering, and multimodal reasoning~\cite{comanici2025_gemini, wang2025_InternVL_3_5, deepseek2025_deepseek_r1_GRPO}.
However, many real-world environments are inherently three-dimensional and can only be observed through multiple partial viewpoints.
This has motivated growing interest in \textit{multi-view 3D visual question answering} (MV3D-VQA)~\cite{hong20233d_3DMV_VQA, zhang2026_think3d}, where a model must integrate observations from multiple viewpoints and reason about spatial relationships within a scene.
Unlike conventional visual question answering that relies on a single image, MV3D-VQA requires aggregating fragmented visual evidence across views to form a coherent understanding of the underlying 3D scene~\cite{hong20233d_3DMV_VQA, fu2024_blink, wang2026understanding_mindcube, zhang2026_think3d}.
Such spatial reasoning is essential for embodied and physical intelligence~\cite{driess2023_palm_embodied, kim2024_openvla, zitkovich2023_rt_vla}, where agents must operate under partial observability and reason about object relations, occlusions, and hypothetical movements.

%

Despite these advances, existing MLLMs still struggle to construct and utilize coherent \textit{three-dimensional spatial representations} from partially observable multi-view inputs~\cite{wang2026understanding_mindcube, fu2024_blink, huang2025_3d_r1, zhang2026_think3d, yang2025_thinking_in_space_vsi_bench}.
Empirical studies on multi-view reasoning benchmarks reveal several systematic failure modes: models frequently produce inconsistent predictions across viewpoints, exhibit unstable reasoning under occlusion, and perform close to random on compositional spatial queries that require integrating evidence from multiple views~\cite{hong20233d_3DMV_VQA, fu2024_blink, wang2026understanding_mindcube, zhang2026_think3d}.
These observations suggest that the core challenge is not merely the availability of visual inputs, but the difficulty of constructing and maintaining a consistent scene-level representation that supports structured spatial reasoning.
Although MLLMs may implicitly encode partial spatial knowledge, they often fail to align, retrieve, and update this information coherently across viewpoints, preventing fragmented observations from being organized into a unified representation of the underlying 3D scene.


Recent work has begun to address this limitation by introducing explicit intermediate representations.
MindCube~\cite{wang2026understanding_mindcube} and related studies~\cite{hong20233d_3DMV_VQA, guo2026_Ego3D_Bench, zhang2026_think3d} demonstrate that providing additional views alone yields limited gains; instead, they show that performance improves when models first construct an intermediate representation such as a \emph{cognitive map} and then reason over it.
These findings establish the \emph{map-then-reason} paradigm as a promising direction: the cognitive map functions as a persistent spatial workspace that organizes fragmented observations into a coherent scene-level structure for downstream tasks such as global scene summarization (spatial memory) and question-conditioned view planning.

%
%

\begin{figure}[t]
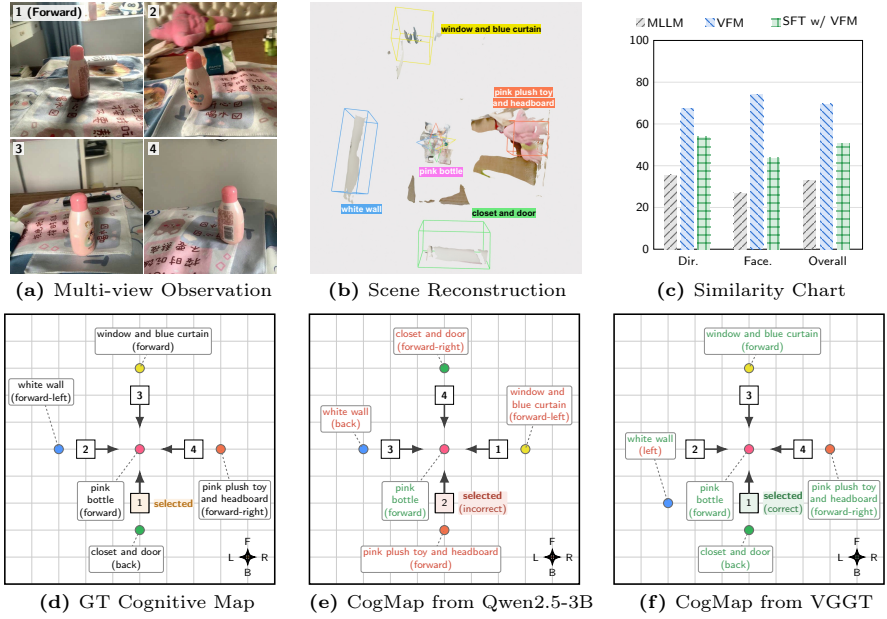

    \centering

    \begin{minipage}{0.98\textwidth}
        \centering

        \newlength{\colgap}
    \newlength{\colw}
    \newlength{\toprowh}
    \newlength{\bottomrowh}
    
    \setlength{\colgap}{0.01\linewidth}
    \setlength{\colw}{\dimexpr(\linewidth-2\colgap)/3\relax}
    
    \setlength{\toprowh}{0.30\textwidth}
    \setlength{\bottomrowh}{0.30\textwidth}
            





        \makebox[\linewidth][c]{%
            \begin{subfigure}[t]{\colw}
                \centering
                \begin{minipage}[t][\toprowh][t]{\linewidth}
                    \centering
                    \adjustbox{max width=\linewidth, max totalheight=\toprowh, keepaspectratio, valign=t}{%
                        \input{figures/intro_figure/subfig_1_multiview.tex}%
                    }
                \end{minipage}
                \caption{Multi-view Observation} 
            \end{subfigure}
            \hfill
            \begin{subfigure}[t]{\colw}
                \centering
                \begin{minipage}[t][\toprowh][t]{\linewidth}
                    \centering
                    \adjustbox{max width=\linewidth, max totalheight=\toprowh, keepaspectratio, valign=t}{%
                        \input{figures/intro_figure/subfig_2_vggt_pointcloud.tex}%
                    }
                \end{minipage}
                \caption{Scene Reconstruction}
            \end{subfigure}
            \hfill
            \begin{subfigure}[t]{\colw}
                \centering
                \begin{minipage}[t][\toprowh][t]{\linewidth}
                    \centering
                    \adjustbox{max width=\linewidth, max totalheight=\toprowh, keepaspectratio, valign=t}{%

\begin{tikzpicture}[font=\sffamily]
    \definecolor{baseCol}{RGB}{120,120,120}
    \definecolor{pseudoCol}{RGB}{58,126,224}
    \definecolor{hybridCol}{RGB}{55,166,92}

    \def\S{3.0}

    \newcommand{\Y}[1]{{\S*(#1)/100}}

    \path[use as bounding box] (-0.35,-0.35) rectangle (\S+0.15,\S+0.55);

    \tikzset{
        tick/.style={font=\fontsize{5.0}{5.0}\selectfont\sffamily, inner sep=0pt},
        metricLabel/.style={font=\fontsize{5.0}{5.0}\selectfont\sffamily, inner sep=0pt},
        legendLabel/.style={font=\fontsize{5.0}{5.0}\selectfont\sffamily, inner sep=0pt}
    }

    \makeatletter
    \pgfdeclarepatternformonly[\pgflinewidth]{dense north east lines}
    {\pgfqpoint{0pt}{0pt}}{\pgfqpoint{3pt}{3pt}}{\pgfqpoint{3pt}{3pt}}{%
      \pgfsetlinewidth{\pgflinewidth}
      \pgfpathmoveto{\pgfqpoint{0pt}{0pt}}
      \pgfpathlineto{\pgfqpoint{3pt}{3pt}}
      \pgfusepath{stroke}
    }
    \pgfdeclarepatternformonly[\pgflinewidth]{dense vertical lines}
    {\pgfqpoint{0pt}{0pt}}{\pgfqpoint{3pt}{3pt}}{\pgfqpoint{3pt}{3pt}}{%
        \pgfsetlinewidth{\pgflinewidth}
        \pgfpathmoveto{\pgfqpoint{0pt}{3pt}}
        \pgfpathlineto{\pgfqpoint{3pt}{0pt}}
        \pgfusepath{stroke}
    }
    \pgfdeclarepatternformonly[\pgflinewidth]{dense grid}
    {\pgfqpoint{0pt}{0pt}}{\pgfqpoint{3pt}{3pt}}{\pgfqpoint{3pt}{3pt}}{%
      \pgfsetlinewidth{\pgflinewidth}
      \pgfpathmoveto{\pgfqpoint{0pt}{1.5pt}}
      \pgfpathlineto{\pgfqpoint{3pt}{1.5pt}}
      \pgfpathmoveto{\pgfqpoint{1.5pt}{0pt}}
      \pgfpathlineto{\pgfqpoint{1.5pt}{3pt}}
      \pgfusepath{stroke}
    }
    \makeatother

    \tikzset{
      patMLLM/.style={pattern=dense north east lines, pattern color=baseCol!95!black, line width=0.45pt},
      patVFM/.style ={pattern=dense vertical lines,   pattern color=pseudoCol!95!black, line width=0.45pt},
      patSFT/.style ={pattern=dense grid,             pattern color=hybridCol!95!black, line width=0.45pt}
    }

    \foreach \v in {20,40,60,80,100} { 
        \draw[gray!25, line width=0.2pt] (0,{ \Y{\v} }) -- (\S,{ \Y{\v} });
    }
    \draw[black, line width=0.35pt] (0,0) rectangle (\S,\S);

    \foreach \v in {0,20,40,60,80,100} {
        \node[tick, anchor=east] at (-0.06,{ \Y{\v} }) {\v};
    }

    \def\gW{0.70}
    \def\bW{0.18}
    \def\gap{0.06}

    \newcommand{\tripbars}[5]{%
        \fill[baseCol!20]   ({#1-0.5*\gW},0) rectangle ++(\bW,{ \Y{#2} });
        \fill[patMLLM]      ({#1-0.5*\gW},0) rectangle ++(\bW,{ \Y{#2} });

        \fill[pseudoCol!20]  ({#1-0.5*\gW+\bW+\gap},0) rectangle ++(\bW,{ \Y{#3} });
        \fill[patVFM]       ({#1-0.5*\gW+\bW+\gap},0) rectangle ++(\bW,{ \Y{#3} });

        \fill[hybridCol!20]  ({#1-0.5*\gW+2*(\bW+\gap)},0) rectangle ++(\bW,{ \Y{#4} });
        \fill[patSFT]       ({#1-0.5*\gW+2*(\bW+\gap)},0) rectangle ++(\bW,{ \Y{#4} });

        \node[metricLabel, anchor=north] at (#1,-0.10) {#5};
    }


    \tripbars{0.50}{35.6}{67.6}{53.8}{Dir.}
    \tripbars{1.50}{27.0}{74.2}{43.7}{Face.}
    \tripbars{2.50}{33.0}{69.6}{50.8}{Overall}

    \begin{scope}[xshift=-10.00] 
        \fill[baseCol!20] (0.10,\S+0.20) rectangle ++(0.12,0.12);
        \fill[patMLLM]    (0.10,\S+0.20) rectangle ++(0.12,0.12);
        \node[legendLabel, anchor=west] at (0.25,\S+0.26) {MLLM};
    
        \fill[pseudoCol!20] (1.10,\S+0.20) rectangle ++(0.12,0.12);
        \fill[patVFM]      (1.10,\S+0.20) rectangle ++(0.12,0.12);
        \node[legendLabel, anchor=west] at (1.25,\S+0.26) {VFM};
    
        \fill[hybridCol!20] (2.05,\S+0.20) rectangle ++(0.12,0.12);
        \fill[patSFT]      (2.05,\S+0.20) rectangle ++(0.12,0.12);
        \node[legendLabel, anchor=west] at (2.20,\S+0.26) {SFT w/ VFM};
    \end{scope}
\end{tikzpicture}%
%
                    }
                \end{minipage}
                \caption{Similarity Chart}
            \end{subfigure}
            \hfill
        }

        \par\smallskip

        \makebox[\linewidth][c]{%
        \begin{subfigure}[t]{\colw}
            \centering
            \begin{minipage}[t][\bottomrowh][t]{\linewidth}
                \centering
                \adjustbox{max width=\linewidth, max totalheight=\bottomrowh, keepaspectratio, valign=t}{%
                    \input{figures/intro_figure/subfig_4_gt_cogmap.tex}%
                }
            \end{minipage}
            \caption{GT Cognitive Map}
            \label{fig:intro:d}
        \end{subfigure}
        \hfill
        \begin{subfigure}[t]{\colw}
            \centering
            \begin{minipage}[t][\bottomrowh][t]{\linewidth}
                \centering
                \adjustbox{max width=\linewidth, max totalheight=\bottomrowh, keepaspectratio, valign=t}{%
                    \input{figures/intro_figure/subfig_5_mllm_cogmap.tex}%
                }
            \end{minipage}
            \caption{CogMap from Qwen2.5-3B}
            \label{fig:intro:e}
        \end{subfigure}
        \hfill
        \begin{subfigure}[t]{\colw}
            \centering
            \begin{minipage}[t][\bottomrowh][t]{\linewidth}
                \centering
                \adjustbox{max width=\linewidth, max totalheight=\bottomrowh, keepaspectratio, valign=t}{%
                    \input{figures/intro_figure/subfig_6_vggt_scene_cogmap.tex}%
                }
            \end{minipage}
            \caption{CogMap from VGGT} \hfill
            \label{fig:intro:f}
        \end{subfigure}%
    }
    \end{minipage}

    \caption{
        \textbf{Cognitive map and metric comparison.}
        Top: multi-view observations, VFM-based scene reconstruction (VGGT+SAM3), and similarity scores to the GT cognitive map.
        The similarity is reported from three aspects: \emph{Dir.} measures inter-object directional relations (relative spatial ordering between objects), \emph{Face.} measures the facing orientation of the viewpoint, and \emph{Overall} is a weighted combination of \emph{Dir.} and \emph{Face.}
        We compute Dir. as pairwise directional agreement and Face. as viewpoint-facing agreement; Overall is their weighted average.
        Bottom: GT cognitive map, an MLLM-generated cognitive map (Qwen2.5-3B), and a VFM-derived cognitive map.
        The VFM-derived cognitive map achieves higher similarity to GT than the MLLM-generated map, suggesting that VGGT+SAM3 can provide a reliable pseudo-structural signal when GT cognitive maps are unavailable.
    }
    \label{fig:intro_cogmap_among_group458}
\end{figure}

Despite this progress, several key challenges remain.
First, most existing approaches rely solely on allocentric (world-centered) representations.
While allocentric maps provide a stable global reference frame, reasoning tasks often involve egocentric queries such as viewpoint-dependent relations or hypothetical movements, creating a representation misalignment between the constructed map and the reasoning process.
Second, MLLMs often struggle to produce geometrically consistent 3D structures even when generating intermediate maps.
In contrast, specialized 3D Vision Foundation Models (VFMs)~\cite{wang2025_vggt, wang2024_dust3r}, such as VGGT, exhibit strong capabilities in multi-view geometric reasoning, including depth estimation, pose consistency, and object-level spatial layout reconstruction.
This reveals a clear representation quality gap between language-driven map generation and geometry-grounded vision models.
As illustrated in~\Cref{fig:intro_cogmap_among_group458}, cognitive maps derived from VGGT~\cite{wang2025_vggt} are more similar to the ground-truth cognitive map than those generated by Qwen2.5-3B~\cite{qwen2025_qwen2_5}, suggesting that VFM-based representations can provide more reliable structural signals for spatial reasoning.
Third, most existing reinforcement learning approaches rely primarily on sparse rewards defined by final answer correctness, which provides limited guidance for learning spatially structured reasoning processes.

%
%

In this work, we propose DR-MV3D (Dense Reward for Multi-View 3D reasoning), a map-grounded multi-view reasoning framework that explicitly addresses these challenges as in ~\Cref{fig:teaser}.
To resolve the allocentric–egocentric representation misalignment (Challenge 1), our approach constructs a global \emph{allocentric cognitive map} summarizing scene-level spatial structure from multi-view observations, and then grounds the acquired evidence into viewpoint-aligned \emph{egocentric representations} that match the reasoning dynamics of MLLMs.
To bridge the geometric reliability gap in language-driven map generation (Challenge 2), we incorporate a 3D Vision Foundation Model (VFM) as a structural prior to regularize cognitive map construction toward geometrically consistent 3D structure.
Conditioned on the global map and the input question, the model further plans a question-aware \emph{view trajectory} that selectively acquires evidence from informative viewpoints.
Finally, to alleviate sparse credit assignment in long-horizon multi-view reasoning (Challenge 3), DR-MV3D optimizes the entire pipeline through trajectory-level policy optimization with a multi-level reward that is \emph{dense along the trajectory} (i.e., step-wise feedback for each view/action)~\cite{chan2024_dense_reward, zhang2025r1vl_StepGRPO}, jointly supervising map consistency, informative view selection, and final answer correctness.
To further improve the geometric reliability of the learned spatial representation, we additionally incorporate supervision from a pretrained 3D Vision Foundation Model.
Specifically, we utilize geometry-consistent scene representations produced by models such as VGGT~\cite{wang2025_vggt} combined with SAM3~\cite{carion2025_sam3} as pseudo-structural targets.
These signals softly align the MLLM-generated cognitive maps with physically plausible 3D layouts without requiring manual annotations.


Our contributions are summarized as follows.
We propose \textbf{DR-MV3D}, a map-grounded multi-view reasoning framework that integrates observations into global allocentric representations and subsequently aligns them with egocentric reasoning dynamics.
To address the limitations of sparse supervision, we introduce a dense reward learning scheme that utilizes pretrained vision foundation models to provide step-wise feedback for both scene consistency and informative viewpoint selection.
Extensive evaluations demonstrate the effectiveness of DR-MV3D, where our 3B model achieves competitive performance across multiple benchmarks, notably reaching 66.5\% accuracy on MindCube-Tiny, a 28.7\%p improvement over the baseline (Qwen2.5-3B~\cite{qwen2025_qwen2_5}), while also delivering superior results on VSI-Bench (37.1 Avg) and BLINK (MV) (56.4).
These findings highlight that dense structural supervision provides a more effective learning signal for multi-step spatial reasoning than standard answer-level optimization.

\section{Related Work}
\label{sec: Related Work}

\textbf{3D Understanding and Spatial Reasoning in VLMs/MLLMs.}
Recent benchmarks and analyses show that, despite strong 2D perception and language-driven reasoning, MLLMs remain brittle in 3D understanding under partial observability and multi-view evidence~\cite{hong20233d_3DMV_VQA, wang2026understanding_mindcube, guo2026_Ego3D_Bench, yang2025_thinking_in_space_vsi_bench, fu2024_blink, linghu2024MSQA}.
Multi-image or multi-view evaluations highlight recurring failures in viewpoint consistency, relative spatial relations, and spatial memory, suggesting that correct answers can stem from shortcuts rather than coherent scene-level representations~\cite{fu2024_blink, yang2025_thinking_in_space_vsi_bench, yang2025mmsi, wang2026understanding_mindcube, guo2026_Ego3D_Bench}.


Prior work addresses these limitations in two complementary ways: (i) injecting 3D-aware representations, such as learning compact 3D scene abstractions from multi-view images, aligning multimodal features with pretrained 3D foundations, or distilling geometric knowledge from 3D vision foundation models into VLMs~\cite{hong20233d_3DMV_VQA, huang20253drs, zhang2026_think3d, lee2025_perspective_aware, lee2025_3d_vlm_gd}, and (ii) improving 3D reasoning behavior via explicit reasoning traces/CoT supervision, perspective-aware mental imagery, or multi-reward RL for long-horizon spatial reasoning and view selection~\cite{linghu2025scenecot, huang2025_3d_r1, lee2025_perspective_aware, liu2025nav-r1, yuan2025embodied-r1}.
However, such supervision is often sparse (answer-level), relies on task-specific intermediate annotations, or distills geometric priors only into static feature representations rather than supervising the multi-step reasoning process itself.
Our work complements these efforts by introducing \emph{dense rewards} that directly shape multi-view spatial reasoning: a global reward for cross-view scene consistency and a local reward for informative viewpoint selection, enabled by scalable pseudo-structural supervision from pretrained vision foundation models.



\textbf{Verifiable Rewards and GRPO for Multimodal Reasoning.}
A growing body of work replaces human preference supervision with verifiable rewards~\cite{deepseek2025_deepseek_r1_GRPO, shao2024_deepseekmath_GRPO,su2025_verifiable_rewards,wen2025_verifiable_rewards}, where learning signals are computed by objective checkers (e.g., exact-match correctness, rule-based constraints, or metric-driven evaluators).
In this paradigm, GRPO~\cite{shao2024_deepseekmath_GRPO} has emerged as a practical reinforcement learning (RL) optimization strategy that avoids an explicit value/critic model by estimating baselines from group-wise samples, enabling scalable RL for reasoning behaviors.
Building on this idea, DeepSeek-R1~\cite{deepseek2025_deepseek_r1_GRPO} demonstrates that strong long-horizon reasoning can be induced primarily from correctness-based, automatically verifiable rewards, reducing dependence on curated chain-of-thought supervision.

Recent extensions bring verifiable-reward RL to vision-language models by designing rewards that evaluate intermediate reasoning steps and grounded perception outputs.
R1-VL~\cite{zhang2025r1vl_StepGRPO} proposes step-wise GRPO with rule-based dense rewards for multi-step multimodal reasoning, while Vision-R1~\cite{zhan2025vision-r1} and Visual-RFT~\cite{liu2025visual-rft} explore vision-guided or metric-driven reinforcement fine-tuning to align visual reasoning and perception without human feedback.
Complementary directions improve generalization via curriculum RL~\cite{deng2025currreft}, two-stage rule-based RL for compact MLLMs~\cite{peng2025lmm-r1}, iterative self-improvement~\cite{deng2025openvlthinker}, and cross-modal formalization for robust multimodal reasoning~\cite{yang2025r1-onevision}.
Related efforts also leverage reasoning-aware training to produce grounded outputs in segmentation and referring tasks~\cite{lai2024lisa, liu2025seg-zero, liu2025visionreasoner}.
Despite these advances, most prior verifiable-reward frameworks predominantly target single-view 2D settings and do not explicitly enforce cross-view spatial consistency under partial observability, which is central to multi-view 3D VQA.

\section{Proposed Method}
\label{sec: Proposed Method}


\subsection{Problem Definition}
\label{subsec:problem-definition}

We consider multi-view 3D visual question answering (MV3D-VQA)~\cite{fu2024_blink, hong20233d_3DMV_VQA, wang2026understanding_mindcube}, where a scene is observed through a finite set of viewpoints.
Formally, given images $\mathcal{I} = \{ I_1, \dots, I_N \}$ captured from different viewpoints of the same scene and a spatial reasoning question $q$, the goal is to predict the correct answer $a^*$.

Unlike conventional 2D VQA~\cite{antol2015_vqa, goyal2017making_vqa}, MV3D-VQA requires the model to actively aggregate fragmented information across space.
We formulate the problem as trajectory-level policy optimization to reflect this active reasoning process~\cite{yao2022_react, wei2022_CoT}, where the model must sequentially determine which observations are necessary to resolve incomplete spatial evidence.
The MLLM induces a structured trajectory $\tau$ that first builds a coarse allocentric cognitive map $\mathcal{C}^{\text{allo}}$, then selects a view trajectory $\mathcal{V} = (v_1, \dots, v_T)$ conditioned on $(\mathcal{C}^{\text{allo}}, q)$.
For each selected view $v_t$, it constructs a corresponding egocentric map $\mathcal{C}^{\text{ego}}_t$, yielding an egocentric map sequence $\mathcal{C}^{\text{ego}}_{1:T}$ that supports continued reasoning and the final answer $\hat{a}$.
The task objective is to maximize expected answer-level reward:
\begin{equation}
    \max_{\theta} \ \mathbb{E}_{\tau \sim \pi_\theta(\cdot \mid \mathcal{I}, q)} \left[ R_{\text{ans}}(\hat{a}(\tau), a^*) \right],
    \quad
    R_{\text{ans}} = \mathbb{I}[\hat{a}(\tau) = a^*],
\end{equation}
where $\hat{a}(\tau)$ is the predicted answer from trajectory $\tau$, and $\mathbb{I}[\cdot]$ is the indicator function.


\subsection{Overall Framework}
\label{subsec:overall-framework}

\begin{figure}[t]
    \centering
    \includegraphics[width=\linewidth]{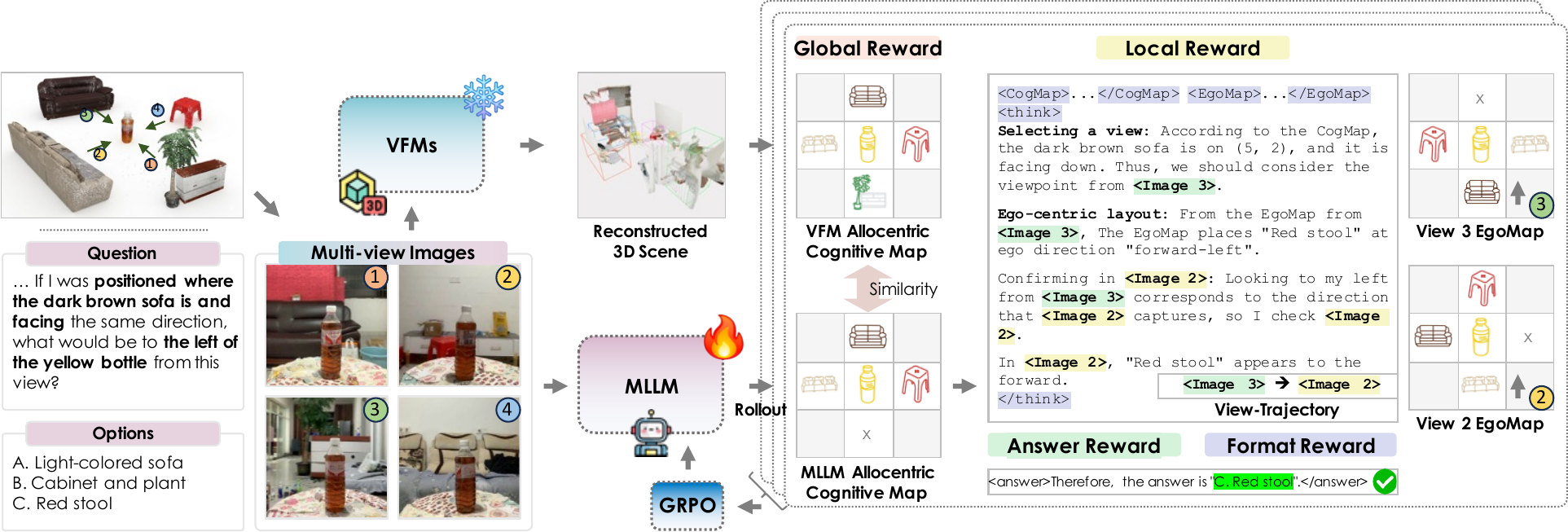}
    \caption{
        \textbf{Overall framework with VFM-guided dense rewards.}
        Given multi-view observations and a question, the MLLM first constructs a \emph{global allocentric cognitive map} and plans a question-conditioned \emph{view trajectory} for evidence acquisition, then grounds the selected views into \emph{egocentric cognitive maps} to predict the final answer.
        During GRPO optimization, we provide multi-level verifiable rewards: \emph{Global Reward} regularizes the MLLM allocentric map toward a geometry-consistent pseudo target produced by frozen VFMs (e.g., VGGT+SAM3); \emph{Local Reward} supervises the step-wise view trajectory by matching planned viewpoints to a reference trajectory; \emph{Answer Reward} scores answer correctness; and \emph{Format Reward} enforces a valid output format.
    }
    
    \label{fig:method-overview}
\end{figure}

Our framework, \textbf{DR-MV3D}, addresses the objective in \Cref{subsec:problem-definition} through a map-grounded multi-view reasoning pipeline for MLLMs.
As illustrated in~\Cref{fig:method-overview}, it decomposes trajectory generation into four stages:
\begin{enumerate}
    \item Global allocentric map construction from multi-view observations,
    \item Question-conditioned view trajectory planning for evidence acquisition,
    \item Egocentric grounding of selected evidence for answer prediction,
    \item Multi-level dense reward supervision for policy optimization.
\end{enumerate}
This subsection summarizes the pipeline; detailed formulations are presented in~\Cref{subsec:global-allocentric-cog-map,subsec:view-trajectory-reasoning,subsec:egocentric-grounding-and-answer-prediction,subsec:reward-design-and-policy-optimization}.
We denote by $\pi_\theta$ the MLLM-induced policy over the full reasoning trajectory, including intermediate cognitive map generation, view selection, egocentric grounding, and answer prediction.

\subsection{Global Allocentric Cognitive Map}
\label{subsec:global-allocentric-cog-map}

Motivated by prior work~\cite{yang2025_thinking_in_space_vsi_bench, wang2026understanding_mindcube, guo2026_Ego3D_Bench}, we construct a global allocentric cognitive map that summarizes scene-level geometry. 
An allocentric representation facilitates a stable, world-centered coordinate system that remains invariant to camera movement, thereby maintaining geometric consistency across disparate viewpoints.
Formally, the global map is defined as:
\begin{equation}
    \mathcal{C}^{\text{allo}} \sim \pi_\theta(\cdot \mid \mathcal{I}).
\end{equation}
where $\pi_\theta$ aggregates multi-view observations into a scene-level representation. 
As an auxiliary geometric prior, we leverage a pretrained and frozen 3D Vision Foundation Model (VFM), instantiated as a combination of VGGT and SAM3~\cite{wang2025_vggt, carion2025_sam3}, to produce a geometry-consistent pseudo ground-truth cognitive map,
\begin{equation}
    \mathcal{C}^{*} = \mathrm{VFMs}(\mathcal{I}) \; _{(\text{e.g., } \mathrm{SAM3}(\mathrm{VGGT}(\mathcal{I})))},
\end{equation}
and supervise the predicted allocentric cognitive map via structural alignment:
\begin{equation}
    R_{\text{global}} = \texttt{sim} \left( {\mathcal{C}}^{\text{allo}}, \mathcal{C}^{*} \right).
\end{equation}
This supervision anchors the learned map to a physically plausible 3D structure without requiring manual annotations.
Further details on the pseudo ground-truth cognitive map construction and the calculation of the similarity function are provided in the supplementary material.


\subsection{View-Trajectory Reasoning}
\label{subsec:view-trajectory-reasoning}

Given the allocentric cognitive map ${\mathcal{C}}^{\text{allo}}$ and the question $q$, the policy generates an ordered local trajectory for evidence acquisition.
Conditioned on $q$, the model prioritizes informative viewpoints while filtering irrelevant visual noise, enabling efficient reasoning in complex 3D scenes.
Formally, the view trajectory is defined as $\mathcal{V} = (v_1, \dots, v_T)$, where each viewpoint is sampled autoregressively as
\begin{equation}
    v_t \sim \pi_\theta(\cdot \mid v_{<t}, {\mathcal{C}}^{\text{allo}}, q).
\end{equation}
This stage performs local view selection grounded in the global allocentric map.
To quantify question-conditioned trajectory quality, we define a step-wise matching reward against a reference trajectory $\mathcal{V}^* = (v_1^*, \dots, v_T^*)$:
\begin{equation}
    R_{\text{local}} = \frac{1}{T} \sum_{t=1}^{T} \mathbb{I}[v_t = v_t^*].
\end{equation}
$R_{\text{local}}$ measures the fraction of viewpoints that match the reference trajectory at the corresponding step, providing direct supervision for ordered view planning.
The reference trajectory $V*$ is deterministically derived from benchmark metadata by identifying the viewpoint that best observes the queried anchor and target objects.


\subsection{Egocentric Grounding and Answer Prediction}
\label{subsec:egocentric-grounding-and-answer-prediction}

After local trajectory planning, the model converts global evidence into egocentric representations aligned with the selected views:
\begin{equation}
    \mathcal{C}^{\text{ego}}_t \sim \pi_\theta(\cdot \mid \mathcal{C}^{\text{allo}}, v_t, \mathcal{I}, q),
\end{equation}
where $\pi_\theta(\cdot)$ performs coordinate rotation, spatial filtering, and reference-frame alignment. 
This conversion is beneficial because MLLMs are primarily pretrained on egocentric (view-aligned) data, and prior work has shown that perspective alignment plays a critical role in visual-spatial problem solving and reasoning~\cite{guo2026_Ego3D_Bench, lee2025_perspective_aware}.
Translating abstract 3D geometry back into the model's native representation, therefore, maximizes reasoning performance.

This conversion ensures that local trajectory planning precedes egocentric grounding and yields the sequence $\mathcal{C}^{\text{ego}}_{1:T}$.
The final answer is predicted from trajectory-conditioned egocentric evidence:
\begin{equation}
    \hat{a}
    =
    \arg\max_{a}
    \pi_\theta\!\left(a \mid \mathcal{V}, \mathcal{C}^{\text{ego}}_{1:T}, \mathcal{I}, q\right).
\end{equation}


\subsection{Reward Design and Policy Optimization}
\label{subsec:reward-design-and-policy-optimization}

The task objective in \Cref{subsec:problem-definition} is answer-level, but optimizing only $R_{\text{ans}}$ yields a sparse learning signal for multi-step spatial reasoning.
Following recent advances in dense reward learning for multi-step reasoning and vision-language policy optimization~\cite{chan2024_dense_reward, zhang2025r1vl_StepGRPO, shao2024_deepseekmath_GRPO}, we therefore optimize the policy with a trajectory-level \emph{dense reward} that provides intermediate supervision aligned with the reasoning stages in~\Cref{subsec:global-allocentric-cog-map,subsec:view-trajectory-reasoning,subsec:egocentric-grounding-and-answer-prediction}.

\subsubsection{Dense Reward Design.}
The reward for trajectory $\tau$ is defined as:
\begin{equation}
    R(\tau) = \lambda_g \cdot R_{\text{global}} + \lambda_l \cdot R_{\text{local}} + \lambda_a \cdot R_{\text{ans}} + \lambda_f \cdot R_{\text{fmt}}.
    \label{eq:dense_reward}
\end{equation}
Each term supervises a different stage of the pipeline. We reuse $R_{\text{global}}$ from \Cref{subsec:global-allocentric-cog-map} and $R_{\text{local}}$ from \Cref{subsec:view-trajectory-reasoning}; the answer-level terms are defined as:
\begin{equation}
\begin{aligned}
    R_{\text{ans}} = \mathbb{I}[\hat{a} = a^*], \quad  R_{\text{fmt}} = \mathbb{I}[\hat{a} \in \mathcal{A}_{\text{valid}}].
\end{aligned}
\end{equation}
Here, $R_{\text{global}}$ and $R_{\text{local}}$ provide intermediate supervision for map quality and trajectory quality, while $(R_{\text{ans}}, R_{\text{fmt}})$ supervise answer correctness and valid output format. In particular, $R_{\text{ans}}$ is the terminal component that directly matches the answer-level task objective.

\subsubsection{Policy Optimization with GRPO.}
We optimize $\pi_\theta$ using Group Relative Policy Optimization (GRPO)~\cite{shao2024_deepseekmath_GRPO, deepseek2025_deepseek_r1_GRPO}.
For each input $(\mathcal{I}, q)$, we sample a group of $K$ trajectories $\{\tau_i\}_{i=1}^{K}$ using the rollout policy $\pi_{\theta_{\mathrm{old}}}$ and compute dense returns $R(\tau_i)$.
We form a standardized group-relative advantage:
\begin{equation}
    \bar{R} \;=\; \frac{1}{K}\sum_{j=1}^{K} R(\tau_j), 
    \qquad
    A_i \;=\; \frac{R(\tau_i)-\bar{R}}{\sqrt{\frac{1}{K}\sum_{j=1}^{K}(R(\tau_j)-\bar{R})^2}+\xi},
    \label{eq:advantages}
\end{equation}
where $\xi$ is a small constant for numerical stability.
Let $r_i(\theta) \;=\; \frac{\pi_\theta(\tau_i \mid \mathcal{I}, q)}{\pi_{\theta_{\mathrm{old}}}(\tau_i \mid \mathcal{I}, q)}$ denote the probability ratio between the current and rollout policies.
The GRPO objective is:
\begin{equation}
\begin{aligned}
    \mathcal{J}_{\text{GRPO}}(\theta)
    &=
    \mathbb{E}_{(\mathcal{I},q)\sim\mathcal{D},\ \tau_{1:K}\sim\pi_{\theta_{\mathrm{old}}}}
    \Bigg[
    \frac{1}{K}\sum_{i=1}^{K}
    \min\!\Big(
    r_i(\theta)A_i,\;
    \operatorname{clip}\!\big(r_i(\theta),\,1-\epsilon,\,1+\epsilon\big)A_i
    \Big)
    \\[-0.2em]
    &\qquad
    -
    \beta\,
    D_{\mathrm{KL}}\!\big(\pi_\theta(\cdot\mid \mathcal{I},q)\,\|\,\pi_{\mathrm{ref}}(\cdot\mid \mathcal{I},q)\big)
    \Bigg].
    \label{eq:GRPO_objective}
\end{aligned}
\end{equation}
We update $\theta$ by maximizing $\mathcal{J}_{\text{GRPO}}(\theta)$ via gradient ascent.
This objective stabilizes policy updates without requiring a separate value network and improves credit assignment across global mapping, view planning, and answer prediction.



\section{Experiments}
\label{sec: Experiments}

\subsection{Experimental Setup}
\label{subsec:experimental-setup}

\subsubsection{Datasets.}
\label{subsubsec:datasets}

We evaluate on three benchmarks: MindCube~\cite{wang2026understanding_mindcube}, VSI-Bench~\cite{yang2025_thinking_in_space_vsi_bench}, and BLINK (MV)~\cite{fu2024_blink}.
MindCube tests view-dependent 3D spatial reasoning and directional consistency.
The dataset consists of 10K training examples and a 1K evaluation set (MindCube-Tiny); throughout the Experiments section, we denote MindCube-Tiny as ``MindCube'' for simplicity.
VSI-Bench focuses on embodied route planning and compositional spatial relations.
BLINK (MV) evaluates multi-view multimodal reasoning under partial and distributed visual evidence.
Additional dataset details (data format, question types, and preprocessing) are provided in the supplementary material.

\subsubsection{Baselines.}
\label{subsubsec:baselines}

We compare against representative multi-image MLLM baselines~\cite{qwen2025_qwen2_5,yang2025_qwen3,singh2025_gpt5,hurst2024_gpt_4o,comanici2025_gemini,zhu2025_internvl3,deng2025_bagel_mot} and prior 3D-aware approaches~\cite{wu2025_spatial_mllm,wang2026understanding_mindcube,cai2025scaling_SenseNova_SI,zhang2026_think3d}.
We keep baseline discussion brief in the main paper and defer prompts, training details, and additional comparisons to the supplementary material.

\subsubsection{Evaluation Metrics.}
\label{subsubsec:evaluation-metrics}

We follow standard benchmark protocols: accuracy for MindCube and BLINK (MV), and per-category scores with an overall average (Avg) for VSI-Bench.

\subsubsection{Implementation Details.}
\label{subsubsec:implementation-details}

DR-MV3D is built on Qwen2.5-VL-3B-Instruct~\cite{qwen2025_qwen2_5}.
We first perform supervised fine-tuning (SFT) to learn the structured 3D reasoning pipeline, and then further optimize the model with GRPO using the proposed dense reward.
In addition, we construct a pseudo-allocentric cognitive map using VGGT~\cite{wang2025_vggt} and SAM3~\cite{carion2025_sam3}, which provides global spatial context for downstream reasoning.
Detailed training configurations (data formatting, optimization hyperparameters, GRPO settings, and map construction details) are provided in the supplementary material.

\newcolumntype{L}{>{\raggedright\arraybackslash}X}
\newcolumntype{C}{>{\centering\arraybackslash}c}

\begin{table}[t]
\centering
\scriptsize

    \caption{
        \textbf{Results on MindCube (MindCube-Tiny).}
        We report accuracy (\%) on the three question categories (\emph{Rotation}, \emph{Among}, \emph{Around}) and the overall score, comparing against proprietary and open-weight multi-image MLLMs as well as prior 3D-aware methods.
        Our model, built on Qwen2.5-VL-3B, consistently improves with SFT and further gains from GRPO with dense rewards, achieving the best overall performance among open-weight baselines.
        Anno. indicates whether ground-truth intermediate cognitive-map/trajectory annotations are used during training.
    }
    \label{tab:mindcube_tiny}

    \resizebox{1.0\textwidth}{!}{
    \begin{threeparttable}
    
        \begin{tabular}{l c c c c c c}
            \toprule
            \multirow{2}{*}{\textbf{Method}} &
            \multirow{2}{*}{\textbf{Params}} &
            \multirow{2}{*}{\textbf{Anno.}} &
            \multicolumn{4}{c}{\textbf{MindCube-Tiny}} \\
            \cmidrule(l){4-7}
            & & & {Rotation} & {Among} & {Around} &\textbf{Overall} \\
            \midrule
            Random (chance)    & -  & \xmark & 38.0 & 37.0 & 32.8 & 34.9 \\
            Random (frequency) & -  & \xmark & 35.7 & 33.3 & 31.8 & 33.0 \\
            \midrule
            \rowcolor{gray!15}
            \multicolumn{7}{@{}l@{}}{\textit{Proprietary Models}} \\
            GPT-4.1~\cite{hurst2024_gpt_4o}                                                  & -  & \xmark & 60.0 & 46.7 & 55.0 & 49.6 \\
            Gemini-2.5-Pro~\cite{comanici2025_gemini}                                        & -  & \xmark & 85.0 & 49.2 & 58.3 & 59.3 \\
            GPT-5~\cite{singh2025_gpt5}                                                      & -  & \xmark & 94.5 & 38.2 & 68.4 & 56.3 \\
            \midrule
            \rowcolor{gray!15} \multicolumn{7}{@{}l@{}}{\textit{Open-Weight Multi Image Models}}                              \\
            Bagel-MoT~\cite{deng2025_bagel_mot}                                              & 7B & \xmark & 34.5 & 31.3 & 42.8 & 34.7 \\
            Qwen2.5-VL-Instruct~\cite{qwen2025_qwen2_5}                                      & 3B & \xmark & 34.0 & 36.0 & 45.2 & 37.8 \\
            Qwen2.5-VL-Instruct~\cite{qwen2025_qwen2_5}                                      & 7B & \xmark & 37.5 & 31.2 & 38.0 & 34.0 \\
            Qwen3-VL-Instruct~\cite{yang2025_qwen3}                                          & 4B & \xmark & 26.0 & 26.7 & 46.0 & 31.1 \\
            Qwen3-VL-Thinking~\cite{yang2025_qwen3}                                          & 4B & \xmark & \textbf{55.0} & 28.7 & 44.8 & 37.5 \\
            InternVL3~\cite{zhu2025_internvl3}                                               & 2B & \xmark & 28.9 & 36.9 & 45.6 & 37.5 \\
            InternVL3~\cite{zhu2025_internvl3}                                               & 8B & \xmark & 36.5 & 38.1 & 53.6 & 41.5 \\
            Spatial-MLLM~\cite{wu2025_spatial_mllm}                                          & 4B & \xmark & 32.5 & 47.5 & 35.0 & 42.8 \\
            MindCube-RawQA-SFT$_\text{(Qwen2.5-VL-Instruct)}$~\cite{wang2026understanding_mindcube}      & 3B & \cmark & 34.0 & 51.0 & 67.6 & 51.7 \\
            MindCube-CGMap-SFT$_\text{(Qwen2.5-VL-Instruct)}$~\cite{wang2026understanding_mindcube}      & 3B & \cmark & 34.5 & 54.2 & 70.8 & 54.4 \\
            MindCube-CGMap-FFR-SFT$_\text{(Qwen2.5-VL-Instruct)}$~\cite{wang2026understanding_mindcube}  & 3B & \cmark & 31.5 & 49.8 & 65.6 & 50.1 \\
            MindCube-CGMap-FFR-RL$_\text{(Qwen2.5-VL-Instruct)}$~\cite{wang2026understanding_mindcube}   & 3B & \cmark & 33.0 & 53.7 & 70.4 & 53.7 \\
            SenseNova-SI$_\text{(Bagel-MoT)}$~\cite{cai2025scaling_SenseNova_SI}             & 7B & \xmark & 37.5 & 57.1 & 46.8 & 50.8 \\
            SenseNova-SI$_\text{(InternVL3)}$~\cite{cai2025scaling_SenseNova_SI}             & 2B & \xmark & 33.5 & 44.4 & 40.0 & 41.2 \\
            Think3D$_\text{(Qwen3-VL-RL)}$~\cite{zhang2026_think3d}                          & 4B & \xmark & 41.7 & 35.0 & 39.2 & 41.7 \\
            \rowcolor{yellow!15} DR-MV3D (Ours)$_\text{(Qwen2.5-VL-Instruct w/ SFT)}$ w/o Anno.                                                            & 3B & \xmark & 36.5 & 52.2 & 64.0 & 53.6 \\
            \rowcolor{yellow!15} DR-MV3D (Ours)$_\text{(Qwen2.5-VL-Instruct w/ SFT)}$                                                                       & 3B & \cmark & 42.0 & 66.3 & 69.2 & 62.4 \\
            \rowcolor{yellow!15} DR-MV3D (Ours)$_\text{(Qwen2.5-VL-Instruct w/ SFT + GRPO)}$ w/o Anno.                                                     & 3B & \xmark & 36.5 & 60.7 & 67.6 & 57.7 \\
            \rowcolor{yellow!15} DR-MV3D (Ours)$_\text{(Qwen2.5-VL-Instruct w/ SFT + GRPO)}$                                                                & 3B & \cmark & 43.0 & \textbf{71.3} & \textbf{73.6} & \textbf{66.5} \\
            \bottomrule
        \end{tabular}
        \end{threeparttable}
    }

\end{table}

\begin{table}[t]
\centering
\caption{
    \textbf{VSI-Bench results.}
    Comparison with multi-image MLLMs and 3D-aware baselines.
}
\label{tab:vsi-bench}

\resizebox{\linewidth}{!}{
    \begin{tabular}{@{}lccccc@{}}
        \toprule
        \textbf{Method} & \textbf{Route Plan} & \textbf{Rel. Dir.} & \textbf{Rel. Dist.} & \textbf{App. Order} & \textbf{Avg} \\
        \midrule

        LLaVA-OneVision-0.5B~\cite{li2024_llavaonevision} 
        & \textbf{34.5} & 36.9 & 28.9 & 5.8 & 26.5 \\

        LLaVA-OneVision-7B~\cite{li2024_llavaonevision} 
        & 29.4 & 35.2 & \textbf{42.5} & 24.4 & 32.9 \\
        
        VILA-1.5-8B~\cite{lin2024_vila} 
        & 31.0 & 34.8 & 32.1 & 24.8 & 30.7 \\

        InternVL2-2B~\cite{internVL2} 
        & 30.4 & 44.1 & 32.1 & 6.3 & 28.2 \\
        
        InternVL2-8B~\cite{internVL2} 
        & 28.9 & 33.4 & 38.0 & \textbf{46.4} & 36.7 \\
        
        Qwen2.5-VL-3B-Instruct~\cite{qwen2025_qwen2_5} 
        & 27.3 & 43.8 & 25.9 & 24.4 & 30.4 \\
        
        \rowcolor{yellow!15}
        DR-MV3D (Ours)$_{\text{(Qwen2.5-VL-3B-Instruct w/ SFT)}}$
        & 30.9 & 43.9 & 35.1 & 26.9 & 34.2 \\
        
        \rowcolor{yellow!15}
        DR-MV3D (Ours)$_{\text{(Qwen2.5-VL-3B-Instruct w/ SFT + GRPO)}}$
        & 32.4 & \textbf{46.6} & 37.8 & 31.4 & \textbf{37.1} \\
        \bottomrule
    \end{tabular}
}
\end{table}

\subsection{Performance Evaluation}
\label{subsec:main-results}

\subsubsection{MindCube.}
We compare DR-MV3D with prior approaches on the MindCube benchmark.
Results in~\Cref{tab:mindcube_tiny} show that our model achieves the best overall accuracy among open-weight multi-image models.
The vanilla Qwen2.5-VL-3B-Instruct baseline scores 37.8 overall, indicating that direct prompting alone is insufficient for reliable multi-view spatial reasoning on this benchmark.
With supervised fine-tuning (SFT), the overall accuracy increases to 62.4, and dense-reward GRPO further improves it to 66.5.
The improvement is most pronounced on \textit{Among} and \textit{Around}, which require compositional reasoning over multiple views and consistent interpretation of relative relations.
Our GRPO model reaches 71.3 on \textit{Among} and 73.6 on \textit{Around}, substantially improving both categories compared to the instruct baseline.
Moreover, our approach surpasses prior map-based baselines trained on the same backbone family, including MindCube-CGMap-SFT (54.4 overall) and MindCube-CGMap-FFR-RL (53.7 overall).
This suggests that dense intermediate supervision provides a more effective learning signal for multi-step spatial reasoning than answer-only optimization, and it improves the model's ability to integrate evidence across views.
This behavior is illustrated in~\Cref{fig:quali_mindcube_1,fig:quali_mindcube_2}, where the baseline misinterprets ego-directions and selects incorrect viewpoints, whereas our model selects an informative viewpoint based on the inferred egocentric map and arrives at the correct answer via step-wise reasoning.
Additional qualitative examples and analyses are provided in the supplementary material.

\subsubsection{VSI-Bench.}
We evaluate sequential route planning from multi-view observations and report four criteria:
Route Plan (route planning), Rel. Dir. (relative direction), Rel. Dist. (relative distance), and App. Order (object appearance order).
As shown in~\Cref{tab:vsi-bench}, our GRPO model achieves the best average score (37.1) among the compared methods.
Our model also attains the comparable results on \textit{Route Plan} (32.4), \textit{Rel. Dir.} (46.6), and \textit{Rel. Dist.} (37.8), which directly measures planning quality and directional reasoning across a sequence of observations.
These gains suggest that dense-reward training transfers beyond single-turn QA and improves navigation-style evaluation, and this requires maintaining a consistent egocentric frame over time.
Relative to the Qwen2.5-VL-3B-Instruct baseline with 30.4, SFT improves the average score to 34.2, and GRPO provides an additional gain to 37.1 while retaining a compact 3B parameter budget.
Overall, the results indicate that our training strategy generalizes to sequential decision-style reasoning from multiple frames beyond static multi-view question answering.

\subsubsection{BLINK (MV).}
We further evaluate transfer to BLINK (MV), which tests distributed multi-view multimodal reasoning.
As shown in~\Cref{tab:blink_mv}, the Qwen2.5-VL-3B baseline scores 42.1, while our SFT model improves to 54.9, and GRPO further increases it to 56.4.
Our GRPO model outperforms large models such as RoBoBrain (55.6) and matches or slightly exceeds Spatial-MLLM (56.0), while also improving over prior RL variants, including Think3D $_\text{(Qwen3-VL-4B-T3RL)}$ (53.4).
Significantly, these gains are obtained with the same 3B backbone, which suggests that the improvement comes from the proposed training signal rather than model scaling.
These outcomes indicate that dense intermediate supervision generalizes beyond MindCube-style evaluation and improves multi-view reasoning when the answer requires fusing evidence across distributed views.

\subsection{Ablation Study}
\label{subsec:ablation-study}

We conduct ablations on MindCube-Tiny to analyze the contribution of (i) the supervision components used in SFT and (ii) the reward design used in GRPO.

\begin{table*}[t]
\centering
\scriptsize
\begin{minipage}[t]{0.52\textwidth}
\centering

\captionof{table}{Results on BLINK (MV).}
\label{tab:blink_mv}

\begin{threeparttable}
\scriptsize
\setlength{\tabcolsep}{4.2pt}
\renewcommand{\arraystretch}{1.06}

\resizebox{\linewidth}{!}{%
\begin{tabular}{l c}
\toprule
\textbf{Method} & \textbf{BLINK (MV)} \\
\midrule
\rowcolor{gray!15}\multicolumn{2}{l}{\textit{Proprietary Models}}\\
GLM-4.5V~\cite{vteam2026glm45}            & 39.9 \\
Doubao-1.5~\cite{Doubao}                  & 50.9 \\
GPT-4.1~\cite{hurst2024_gpt_4o}           & 36.8 \\
Gemini-2.5-Pro~\cite{comanici2025_gemini} & 44.9 \\
\midrule
\rowcolor{gray!15}\multicolumn{2}{l}{\textit{Open-Weight Models}}\\
RoBoBrain~\cite{RoBoBrain}                & 55.6 \\
Spatial-MLLM~\cite{wu2025_spatial_mllm}   & 56.0 \\
VLM-3R~\cite{VLM_3R}                      & 41.4 \\
REVPT~\cite{REVPT}                        & 51.9 \\
Qwen2.5-VL-3B~\cite{qwen2025_qwen2_5}     & 42.1 \\
Qwen3-VL-4B~\cite{yang2025_qwen3}         & 47.9 \\
Qwen3-VL-4B-GRPO~\cite{yang2025_qwen3}    & 52.4 \\
Qwen3-VL-4B-T3RL~\cite{yang2025_qwen3}                       & 46.1 \\
Think3D$_\text{(Qwen3-VL-4B)}$~\cite{zhang2026_think3d}    & 48.6 \\
Think3D$_\text{(Qwen3-VL-4B-T3RL)}$~\cite{zhang2026_think3d} & 53.4 \\
\rowcolor{yellow!15}
DR-MV3D (Ours)$_\text{(Qwen2.5-VL-3B-Instruct w/ SFT)}$      & 54.9 \\
\rowcolor{yellow!15}
DR-MV3D (Ours)$_\text{(Qwen2.5-VL-3B-Instruct w/ SFT + GRPO)}$ & \textbf{56.4} \\
\bottomrule
\end{tabular}%
}
\end{threeparttable}

\end{minipage}
\hfill
\resizebox{0.47\linewidth}{!}{%
\begin{minipage}[t]{0.485\textwidth}
\centering

\begin{threeparttable}
\scriptsize
\setlength{\tabcolsep}{2.6pt}
\renewcommand{\arraystretch}{1.06}

\captionof{table}{
    Ablation of components in SFT.
}
\label{tab:ablation_sft_top}

\begin{tabularx}{\linewidth}{@{}
  >{\centering\arraybackslash}p{0.22\linewidth}
  >{\centering\arraybackslash}p{0.22\linewidth}
  >{\centering\arraybackslash}p{0.22\linewidth}
  >{\centering\arraybackslash}X
@{}}
\toprule
\textbf{Allo. Map} & \textbf{Traj.} & \textbf{Ego. Map} & \textbf{Overall} \\
\midrule
\rowcolor{gray!15}\multicolumn{4}{@{}l}{\textit{w/o SFT (Vanilla Qwen2.5-VL-3B-Instruct)}}\\
\xmark & \xmark & \xmark & 37.8 \\
\rowcolor{gray!15}\multicolumn{4}{@{}l}{\textit{w/ SFT}}\\
\cmark & \xmark & \xmark & 52.8 \\
\xmark & \cmark & \xmark & 58.2 \\
\xmark & \xmark & \cmark & 52.2 \\
\cmark & \xmark & \cmark & 53.6 \\
\cmark & \cmark & \cmark & \textbf{62.4} \\
\bottomrule
\end{tabularx}
\end{threeparttable}

\par\smallskip

\captionof{table}{
    Ablation on rewards in GRPO.
}
\label{tab:ablation_sft_bottom}

\begin{threeparttable}
\scriptsize
\setlength{\tabcolsep}{2.6pt}
\renewcommand{\arraystretch}{1.06}

\begin{tabularx}{\linewidth}{@{}
  >{\centering\arraybackslash}p{0.22\linewidth}
  >{\centering\arraybackslash}p{0.22\linewidth}
  >{\centering\arraybackslash}p{0.22\linewidth}
  >{\centering\arraybackslash}X
@{}}
\toprule
\textbf{Answer \& format} & \textbf{Global} & \textbf{Local} & \textbf{Overall} \\
\midrule
\cmark & \xmark & \xmark & 63.8 \\
\cmark & \cmark & \xmark & 64.9 \\
\cmark & \cmark & \cmark & \textbf{66.5} \\
\bottomrule
\end{tabularx}
\end{threeparttable}

\end{minipage}
}

\end{table*}

\newcommand{\mvimg}[2]{%
    \begin{tikzpicture}
        \node[inner sep=0, outer sep=0] (img)
        {\includegraphics[width=\linewidth,clip,trim={0 85 0 85}]{#1}};
        \path[use as bounding box] (img.south west) rectangle (img.north east);
        \node[
            anchor=north west,
            fill=white,
            fill opacity=0.88,
            text opacity=1,
            inner sep=0.7pt,
            font=\fontsize{17.0}{19.0}\selectfont\sffamily\bfseries
        ] at ([xshift=1pt,yshift=-1pt]img.north west) {#2};
    \end{tikzpicture}%
}

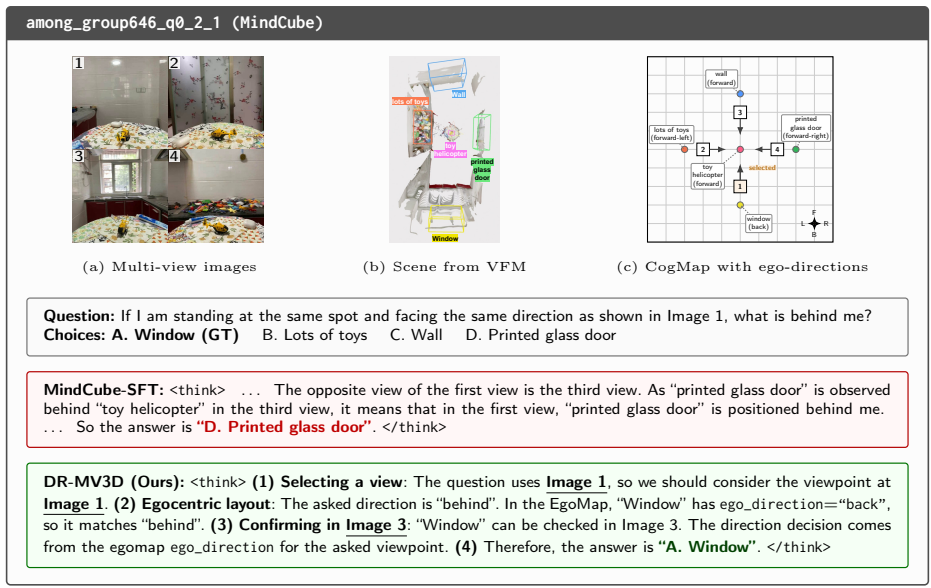
\begin{figure}[t]
    \centering

    \newlength{\qualTopA}
    \newlength{\qualTopB}
    \newlength{\qualTopC}
    \newlength{\qualTopH}
    \newlength{\qualCapH}
    \newlength{\qualImgH}

    \setlength{\qualTopA}{0.31\linewidth}
    \setlength{\qualTopB}{0.27\linewidth}
    \setlength{\qualTopC}{0.36\linewidth}
    \setlength{\qualTopH}{0.29\textwidth}  
    \setlength{\qualTopH}{0.25\textwidth}  
    \setlength{\qualCapH}{1.8em}
    \setlength{\qualImgH}{\dimexpr\qualTopH-\qualCapH\relax}

    \begin{tcolorbox}[
        width=\linewidth,
        colback=black!1,
        colframe=black!70,
        boxrule=0.6pt,
        arc=1.2pt,
        after skip=3pt,
        left=1.5mm,right=1.5mm,top=1.1mm,bottom=1.1mm,
        title=\textcolor{white}{\texttt{\textbf{\scriptsize among\_group646\_q0\_2\_1 (MindCube)}}},
        fonttitle=\sffamily\bfseries,
        coltitle=black
    ]

        \makebox[\linewidth][c]{%
            \begin{minipage}[t][\qualTopH][t]{\qualTopA}
            \centering
        
            \begin{minipage}[c][\qualImgH][c]{\linewidth}
                \centering
                \setlength{\tabcolsep}{0pt}
                \renewcommand{\arraystretch}{0}
                \adjustbox{max width=0.98\linewidth, max totalheight=\qualImgH, keepaspectratio}{%
                    \begin{tabular}{@{}cc@{}}
                        \mvimg{}{1} &  
                        \mvimg{}{2} \\
                        \mvimg{}{3} &
                        \mvimg{}{4}
                    \end{tabular}
                }
            \end{minipage}
        
            \begin{minipage}[b][\qualCapH][c]{\linewidth}
                \centering
                {\tiny (a) Multi-view images}
            \end{minipage}
        \end{minipage}\hfill

            \begin{minipage}[t][\qualTopH][t]{\qualTopB}
                \centering
                \begin{minipage}[c][\qualImgH][c]{\linewidth}
                    \centering
                    \adjustbox{max width=0.98\linewidth, max totalheight=\qualImgH, keepaspectratio}{%
                        \includegraphics[trim={6 6 6 6},clip]{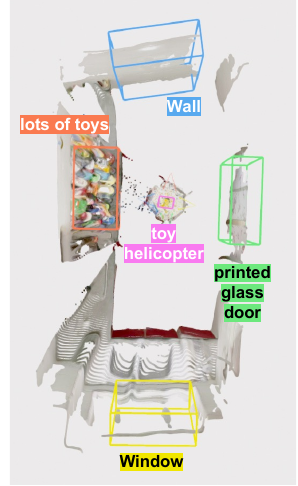}%
                    }
                \end{minipage}

                \begin{minipage}[b][\qualCapH][c]{\linewidth}
                    \centering
                    {\tiny (b) Scene from VFM}
                \end{minipage}
            \end{minipage}\hfill
            \begin{minipage}[t][\qualTopH][t]{\qualTopC}
                \centering
                \begin{minipage}[c][\qualImgH][c]{\linewidth}
                    \centering
                    \adjustbox{max width=0.97\linewidth, max totalheight=\qualImgH, keepaspectratio}{%
\begin{tikzpicture}[font=\fontsize{5.4}{6.0}\selectfont\sffamily]
  \begin{scope}[x=0.46cm,y=-0.46cm]
    \foreach \x in {0,...,10} {\draw[gray!40,line width=0.28pt] (\x,0) -- (\x,10);}
    \foreach \y in {0,...,10} {\draw[gray!40,line width=0.28pt] (0,\y) -- (10,\y);}
    \draw[black,line width=0.75pt] (0,0) rectangle (10,10);

    \definecolor{selView}{RGB}{226,140,28} 

    \definecolor{colToy}{RGB}{255, 92, 135}      
    \definecolor{colWindow}{RGB}{235, 225, 48}   
    \definecolor{colToys}{RGB}{241, 113, 75}     
    \definecolor{colWall}{RGB}{80, 150, 255}     
    \definecolor{colDoor}{RGB}{50, 180, 90}      

    \tikzset{
      objdot/.style={draw=black!60,line width=0.35pt,circle,minimum size=4.4pt,inner sep=0pt},
      objlabel/.style={draw=black!45,fill=white,rounded corners=1.0pt,inner sep=1.0pt,align=center},
      callout/.style={black!55,line width=0.34pt,dash pattern=on 1pt off 1pt},
      viewone/.style={draw=black, fill=white, line width=0.45pt},
      viewselect/.style={draw=black, fill=selView!12, line width=0.45pt},
      viewnum/.style={font=\bfseries\fontsize{5.0}{5.0}\selectfont\sffamily}
    }

    \node[objdot,fill=colToy]    (objToy)    at (5,5) {};
    \node[objdot,fill=colWindow] (objWindow) at (5,8) {};
    \node[objdot,fill=colToys]   (objToys)   at (2,5) {};
    \node[objdot,fill=colWall]   (objWall)   at (5,2) {};
    \node[objdot,fill=colDoor]   (objDoor)   at (8,5) {};

    \coordinate (labToyPos)    at (3.25,6.45);
    \coordinate (labWindowPos) at (6.00,9.00);
    \coordinate (labToysPos)   at (1.30,4.20);
    \coordinate (labWallPos)   at (4.00,1.20);
    \coordinate (labDoorPos)   at (8.55,3.85);

    \draw[callout,shorten <=0.6pt,shorten >=0.8pt] (labToyPos)    -- (objToy.center);
    \draw[callout,shorten <=0.6pt,shorten >=0.8pt] (labWindowPos) -- (objWindow.center);
    \draw[callout,shorten <=0.6pt,shorten >=0.8pt] (labToysPos)   -- (objToys.center);
    \draw[callout,shorten <=0.6pt,shorten >=0.8pt] (labWallPos)   -- (objWall.center);
    \draw[callout,shorten <=0.6pt,shorten >=0.8pt] (labDoorPos)   -- (objDoor.center);

    \node[objlabel] at (labToyPos)    {toy\\helicopter\\(forward)};
    \node[objlabel] at (labWindowPos) {window\\(back)};
    \node[objlabel] at (labToysPos)   {lots of toys\\(forward-left)};
    \node[objlabel] at (labWallPos)   {wall\\(forward)};
    \node[objlabel] at (labDoorPos)   {printed\\glass door\\(forward-right)};

    \def\boxSq{0.34}

    \path[viewselect] (5-\boxSq,7-\boxSq) rectangle (5+\boxSq,7+\boxSq);
    \node[viewnum] at (5,7) {1};
    \node[fill=selView!12,text=selView!80!black,rounded corners=0.6pt,inner sep=0.25pt,anchor=west]
      at (5.45,6.00) {\textbf{selected}};
    \draw[-{Latex[length=2.25mm]},black!75,line width=0.95pt]
      (5,7-\boxSq) -- (5,5.75);

    \path[viewone] (3-\boxSq,5-\boxSq) rectangle (3+\boxSq,5+\boxSq);
    \node[viewnum] at (3,5) {2};
    \draw[-{Latex[length=2.05mm]},black!75,line width=0.75pt]
      (3+\boxSq,5) -- (4.25,5);

    \path[viewone] (5-\boxSq,3-\boxSq) rectangle (5+\boxSq,3+\boxSq);
    \node[viewnum] at (5,3) {3};
    \draw[-{Latex[length=2.05mm]},black!75,line width=0.75pt]
      (5,3+\boxSq) -- (5,4.25);

    \path[viewone] (7-\boxSq,5-\boxSq) rectangle (7+\boxSq,5+\boxSq);
    \node[viewnum] at (7,5) {4};
    \draw[-{Latex[length=2.05mm]},black!75,line width=0.75pt]
      (7-\boxSq,5) -- (5.75,5);

    \coordinate (egoC) at (9.00,9.00);
    \fill[selView] (egoC) circle (0.05);
    \draw[-{Latex[length=1.4mm]},line width=0.55pt] (egoC) -- ++(0.36,0);
    \draw[-{Latex[length=1.4mm]},line width=0.55pt] (egoC) -- ++(-0.36,0);
    \draw[-{Latex[length=1.4mm]},line width=0.55pt] (egoC) -- ++(0,-0.36);
    \draw[-{Latex[length=1.4mm]},line width=0.55pt] (egoC) -- ++(0,0.36);
    \node[anchor=west]  at ($(egoC)+(0.25,0)$) {R};
    \node[anchor=east]  at ($(egoC)+(-0.25,0)$) {L};
    \node[anchor=south] at ($(egoC)+(0,-0.2)$) {F};
    \node[anchor=north] at ($(egoC)+(0,0.2)$) {B};

  \end{scope}
\end{tikzpicture}%
                    }
                \end{minipage}

                \begin{minipage}[b][\qualCapH][c]{\linewidth}
                    \centering
                    {\tiny (c) CogMap with ego-directions}
                \end{minipage}
            \end{minipage}%
        }

        \par\smallskip

        \begin{tcolorbox}[
            width=\linewidth,
            colback=black!2,
            colframe=black!55,
            boxrule=0.45pt,
            arc=0.9pt,
            before skip=1pt,
            after skip=5pt,
            left=1.1mm,right=1.1mm,top=0.5mm,bottom=0.5mm,
            fontupper=\fontsize{6}{7}\selectfont\sffamily
        ]
        \textbf{Question:} If I am standing at the same spot and facing the same direction as shown in Image~1, what is behind me? \\
        \textbf{Choices:} \textbf{A. Window (GT)} \quad B. Lots of toys \quad C. Wall \quad D. Printed glass door
        \end{tcolorbox}

        \begin{tcolorbox}[
            width=\linewidth,
            colback=red!3,
            colframe=red!70!black,
            boxrule=0.45pt,
            arc=0.9pt,
            left=1.1mm,right=1.1mm,top=0.5mm,bottom=0.5mm,
            fontupper=\fontsize{6}{7}\selectfont\sffamily
        ]
        \textbf{MindCube-SFT:} \texttt{<think>} \; \dots \; The opposite view of the first view is the third view. As ``printed glass door'' is observed behind ``toy helicopter'' in the third view, it means that in the first view, ``printed glass door'' is positioned behind me. \; \dots \; So the answer is \textbf{\textcolor{red!75!black}{``D. Printed glass door''}}. \texttt{</think>}
        \end{tcolorbox}

        \begin{tcolorbox}[
            width=\linewidth,
            colback=green!4,
            colframe=green!45!black,
            boxrule=0.45pt,
            arc=0.9pt,
            left=1.1mm,right=1.1mm,top=0.5mm,bottom=0.5mm,
            fontupper=\fontsize{6}{7}\selectfont\sffamily
        ]
        \textbf{DR-MV3D (Ours):} \texttt{<think>} \textbf{(1) Selecting a view}: The question uses \textbf{\underline{Image 1}}, so we should consider the viewpoint at \textbf{\underline{Image 1}}. \textbf{(2) Egocentric layout}: The asked direction is ``behind''. In the EgoMap, ``Window'' has \texttt{ego\_direction}$=$\texttt{``back''}, so it matches ``behind''. \textbf{(3) Confirming in \underline{Image 3}}: ``Window'' can be checked in Image 3. The direction decision comes from the egomap \texttt{ego\_direction} for the asked viewpoint. \textbf{(4)} Therefore, the answer is \textbf{\textcolor{green!25!black}{``A. Window''}}. \texttt{</think>}
        \end{tcolorbox}

    \end{tcolorbox}

    \caption{
        \textbf{Qualitative comparison (Among).}
        Multi-view observations and cognitive-map-based reasoning for an \textit{Among} query (MindCube).
        The baseline confuses egocentric directions and predicts an incorrect object, whereas our method follows the ego-direction cues in the predicted cognitive map step by step and selects the correct answer.
    }
    \label{fig:quali_mindcube_1}
\end{figure}

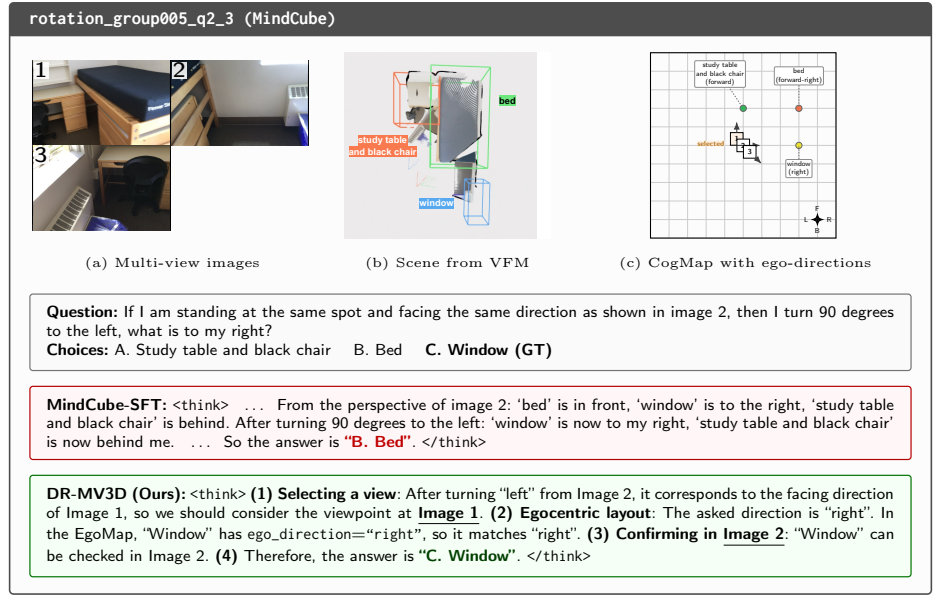
\begin{figure}[t]
    \centering


    \setlength{\qualTopA}{0.31\linewidth}
    \setlength{\qualTopB}{0.27\linewidth}
    \setlength{\qualTopC}{0.36\linewidth}
    \setlength{\qualTopH}{0.29\textwidth}  
    \setlength{\qualTopH}{0.25\textwidth}  
    \setlength{\qualCapH}{1.8em}
    \setlength{\qualImgH}{\dimexpr\qualTopH-\qualCapH\relax}

    \begin{tcolorbox}[
        width=\linewidth,
        colback=black!1,
        colframe=black!70,
        boxrule=0.6pt,
        arc=1.2pt,
        before skip=0pt,
        after skip=3pt,
        left=1.5mm,right=1.5mm,top=1.1mm,bottom=1.1mm,
        title=\textcolor{white}{\texttt{\textbf{\scriptsize rotation\_group005\_q2\_3 (MindCube)}}},
        fonttitle=\sffamily\bfseries,
        coltitle=black
    ]

        \makebox[\linewidth][c]{%
            \begin{minipage}[t][\qualTopH][t]{\qualTopA}
            \centering
        
            \begin{minipage}[c][\qualImgH][c]{\linewidth}
                \centering
                \setlength{\tabcolsep}{0pt}
                \renewcommand{\arraystretch}{0}
                \adjustbox{max width=0.98\linewidth, max totalheight=\qualImgH, keepaspectratio}{%
                    \begin{tabular}{@{}cc@{}}
                        \mvimg{}{1} &
                        \mvimg{}{2} \\  
                        \mvimg{}{3}
                    \end{tabular}
                }
            \end{minipage}
        
            \begin{minipage}[b][\qualCapH][c]{\linewidth}
                \centering
                {\tiny (a) Multi-view images}
            \end{minipage}
        \end{minipage}\hfill

            \begin{minipage}[t][\qualTopH][t]{\qualTopB}
                \centering
                \begin{minipage}[c][\qualImgH][c]{\linewidth}
                    \centering
                    \adjustbox{max width=0.98\linewidth, max totalheight=\qualImgH, keepaspectratio}{%
                        \includegraphics[trim={6 6 6 6},clip]{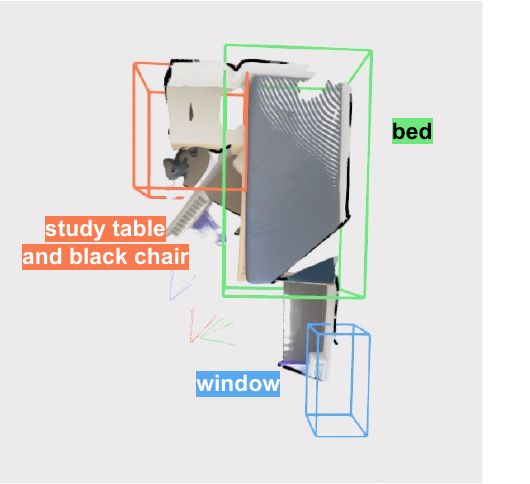}%
                    }
                \end{minipage}

                \begin{minipage}[b][\qualCapH][c]{\linewidth}
                    \centering
                    {\tiny (b) Scene from VFM}
                \end{minipage}
            \end{minipage}\hfill
            \begin{minipage}[t][\qualTopH][t]{\qualTopC}
                \centering
                \begin{minipage}[c][\qualImgH][c]{\linewidth}
                    \centering
                    \adjustbox{max width=0.97\linewidth, max totalheight=\qualImgH, keepaspectratio}{%
\begin{tikzpicture}[font=\fontsize{5.4}{6.0}\selectfont\sffamily]
  \begin{scope}[x=0.46cm,y=-0.46cm]
    \foreach \x in {0,...,10} {\draw[gray!40,line width=0.28pt] (\x,0) -- (\x,10);}
    \foreach \y in {0,...,10} {\draw[gray!40,line width=0.28pt] (0,\y) -- (10,\y);}
    \draw[black,line width=0.75pt] (0,0) rectangle (10,10);

    \definecolor{selView}{RGB}{226,140,28}

    \tikzset{
      objdot/.style={draw=black!60,line width=0.35pt,circle,minimum size=4.4pt,inner sep=0pt},
      objlabel/.style={draw=black!45,fill=white,rounded corners=1.0pt,inner sep=1.0pt,align=center},
      callout/.style={black!55,line width=0.34pt,dash pattern=on 1pt off 1pt},
      innerwhite/.style={fill=white, draw=black, line width=0.25pt},
      viewone/.style={draw=black, fill=white, line width=0.45pt},
      viewnum/.style={font=\bfseries\fontsize{5.0}{5.0}\selectfont\sffamily}
    }

    \definecolor{colBed}{RGB}{241, 113, 75}       
    \definecolor{colWindow}{RGB}{235, 225, 48}    
    \definecolor{colDesk}{RGB}{50, 180, 90}       

    \node[objdot,fill=colBed]    (objBed)  at (8,3) {};
    \node[objdot,fill=colWindow] (objWin)  at (8,5) {};
    \node[objdot,fill=colDesk]   (objDesk) at (5,3) {};

    
    \coordinate (labBedPos)  at (8.00,1.25);
    \coordinate (labWinPos)  at (8.00,6.25);
    \coordinate (labDeskPos) at (3.75,1.15);

    \draw[callout,shorten <=0.6pt,shorten >=0.8pt] (labBedPos)  -- (objBed.center);
    \draw[callout,shorten <=0.6pt,shorten >=0.8pt] (labWinPos)  -- (objWin.center);
    \draw[callout,shorten <=0.6pt,shorten >=0.8pt] (labDeskPos) -- (objDesk.center);

    \node[objlabel] at (labBedPos)  {bed\\(forward-right)};
    \node[objlabel] at (labWinPos)  {window\\(right)};
    \node[objlabel] at (labDeskPos) {study table\\and black chair\\(forward)};

    \def\boxSq{0.34}

    \coordinate (vC) at (5,5);

    \coordinate (v1) at ($(vC)+(-0.35,-0.35)$); 
    \coordinate (v2) at (vC);                   
    \coordinate (v3) at ($(vC)+(+0.35,+0.35)$);   

    \path[viewone, fill=selView!12,text=selView!80!black,] ($(v1)+(-\boxSq,-\boxSq)$) rectangle ($(v1)+(\boxSq,\boxSq)$);
    \node[viewnum] at (v1) {1};
    \draw[-{Latex[length=2.05mm]},black!75,line width=0.75pt]
      ($(v1)+(0,-\boxSq)$) -- ($(v1)+(0,-0.95)$);
    \node[fill=selView!12,text=selView!80!black,rounded corners=0.6pt,inner sep=0.25pt,anchor=west,align=center] at (2.50,4.90) {\textbf{selected}};

    \path[viewone] ($(v2)+(-\boxSq,-\boxSq)$) rectangle ($(v2)+(\boxSq,\boxSq)$);
    \node[viewnum] at (v2) {2};
    \draw[-{Latex[length=2.05mm]},black!75,line width=0.75pt]
      ($(v2)+(-\boxSq,0)$) -- ($(v2)+(+0.95,0)$);

    \path[viewone] ($(v3)+(-\boxSq,-\boxSq)$) rectangle ($(v3)+(\boxSq,\boxSq)$);
    \node[viewnum] at (v3) {3};
    \draw[-{Latex[length=2.05mm]},black!75,line width=0.75pt]
      ($(v3)+(+\boxSq,+\boxSq)$) -- ($(v3)+(+0.65,+0.65)$);

    \coordinate (egoC) at (9.00,9.00);
    \fill[selView] (egoC) circle (0.05);
    \draw[-{Latex[length=1.4mm]},line width=0.55pt] (egoC) -- ++(0.36,0);
    \draw[-{Latex[length=1.4mm]},line width=0.55pt] (egoC) -- ++(-0.36,0);
    \draw[-{Latex[length=1.4mm]},line width=0.55pt] (egoC) -- ++(0,-0.36);
    \draw[-{Latex[length=1.4mm]},line width=0.55pt] (egoC) -- ++(0,0.36);
    \node[anchor=west]  at ($(egoC)+(0.25,0)$) {R};
    \node[anchor=east]  at ($(egoC)+(-0.25,0)$) {L};
    \node[anchor=south] at ($(egoC)+(0,-0.2)$) {F};
    \node[anchor=north] at ($(egoC)+(0,0.2)$) {B};

  \end{scope}
\end{tikzpicture}%
                    }
                \end{minipage}

                \begin{minipage}[b][\qualCapH][c]{\linewidth}
                    \centering
                    {\tiny (c) CogMap with ego-directions}
                \end{minipage}
            \end{minipage}%
        }

        \par\smallskip

        \begin{tcolorbox}[
            width=\linewidth,
            colback=black!2,
            colframe=black!55,
            boxrule=0.45pt,
            arc=0.9pt,
            before skip=1pt,
            after skip=5pt,
            left=1.1mm,right=1.1mm,top=0.5mm,bottom=0.5mm,
            fontupper=\fontsize{6}{7}\selectfont\sffamily
        ]
        \textbf{Question:} If I am standing at the same spot and facing the same direction as shown in image 2, then I turn 90 degrees to the left, what is to my right? \\
        \textbf{Choices:} A. Study table and black chair \quad B. Bed \quad \textbf{C. Window (GT)}
        \end{tcolorbox}

        \begin{tcolorbox}[
            width=\linewidth,
            colback=red!3,
            colframe=red!70!black,
            boxrule=0.45pt,
            arc=0.9pt,
            left=1.1mm,right=1.1mm,top=0.5mm,bottom=0.5mm,
            fontupper=\fontsize{6}{7}\selectfont\sffamily
        ]
        \textbf{MindCube-SFT:} \texttt{<think>} \; \dots \; From the perspective of image 2: `bed' is in front, `window' is to the right, `study table and black chair' is behind. After turning 90 degrees to the left: `window' is now to my right, `study table and black chair' is now behind me. \; \dots \; So the answer is \textbf{\textcolor{red!75!black}{``B. Bed''}}. \texttt{</think>}
        \end{tcolorbox}

        \begin{tcolorbox}[
            width=\linewidth,
            colback=green!4,
            colframe=green!45!black,
            boxrule=0.45pt,
            arc=0.9pt,
            left=1.1mm,right=1.1mm,top=0.5mm,bottom=0.5mm,
            fontupper=\fontsize{6}{7}\selectfont\sffamily
        ]
        \textbf{DR-MV3D (Ours):} \texttt{<think>} \textbf{(1) Selecting a view}: After turning ``left'' from Image 2, it corresponds to the facing direction of Image 1, so we should consider the viewpoint at \textbf{\underline{Image 1}}. \textbf{(2) Egocentric layout}: The asked direction is ``right''. In the EgoMap, ``Window'' has \texttt{ego\_direction}$=$\texttt{``right''}, so it matches ``right''. \textbf{(3) Confirming in \underline{Image 2}}: ``Window'' can be checked in Image 2. \textbf{(4)} Therefore, the answer is \textbf{\textcolor{green!25!black}{``C. Window''}}. \texttt{</think>}
        \end{tcolorbox}

    \end{tcolorbox}
    \caption{
        \textbf{Qualitative comparison (Rotation).}
        Multi-view observations and cognitive-map-based reasoning for a \textit{Rotation} query (MindCube).
        The baseline fails to track the rotated egocentric frame and selects the wrong object, whereas our method maintains consistent ego-direction reasoning under rotation and predicts the correct answer.
    }
    \label{fig:quali_mindcube_2}
\end{figure}

\subsubsection{Component ablation in SFT.}
As shown in~\Cref{tab:ablation_sft_top}, intermediate supervision is most effective during SFT when multiple components are combined.
The vanilla Qwen2.5-3B baseline without SFT achieves 37.8 overall.
When we enable a single component, supervising the allocentric map reaches 52.8, supervising the egocentric map reaches 52.2, and supervising only the trajectory reaches 58.2.
Supervising both maps improves the score to 53.6, suggesting that the two map representations provide complementary supervision but are not sufficient on their own.
Furthermore, adding trajectory supervision for local-view selection yields the best performance of 62.4.
This ablation indicates that strong performance benefits from jointly constraining global spatial structure and step-wise reasoning signals, rather than relying on a single intermediate target.

\subsubsection{Reward ablation in GRPO.}
We also study the effect of reward design during GRPO, as summarized in~\Cref{tab:ablation_sft_bottom}, starting from an SFT model that already produces valid answers and formats.
Using only the answer-and-format reward yields 63.8 overall.
Adding the global reward improves performance to 64.9, and incorporating both global and local rewards achieves the best result of 66.5.
These consistent gains suggest that answer-level optimization alone does not fully capture multi-step spatial reasoning quality, while dense intermediate rewards provide complementary improvements.

\section{Conclusion}
We presented \textbf{DR-MV3D}, a dense-reward learning framework for multi-view 3D visual question answering (MV3D-VQA) under partial observability.
Our method formulates MV3D-VQA as trajectory-level policy optimization: the agent constructs an allocentric cognitive map to maintain global scene coherence, selects question-conditioned viewpoint trajectories to acquire informative evidence, and aligns observations into egocentric maps for step-wise reasoning and answer prediction.
To make these intermediate processes learnable, we introduce dense rewards that directly supervise cross-view consistency and viewpoint selection using pseudo-structural signals from vision foundation models.
Experiments on MindCube, VSI-Bench, and BLINK show consistent gains over strong supervised baselines, highlighting the value of process-level supervision for 3D reasoning.

\newpage

\section*{Acknowledgments}



This work was supported by the Basic Science Research Program through the National Research Foundation of Korea (NRF) funded by the Korea government (MSIT) (No. RS-2025-00520207); Institute of Information \& Communications Technology Planning \& Evaluation (IITP) grants funded by the Korea government (MSIT) (Nos. RS-2024-00457882, 2022-0-01045, 2022-0-00680, RS-2019-II190075); the Advanced GPU Utilization Support Program funded by the Government of the Republic of Korea (Ministry of Science and ICT); the AI Computing Infrastructure Enhancement (GPU Rental Support) User Support Program funded by the Ministry of Science and ICT (MSIT), Republic of Korea (No. RQT-25-120217); a grant partly supported by both IITP (MSIT) and Korea Evaluation Institute of Industrial Technology (KEIT) (MOTIE) (No. RS-2025-02217259); and the next-generation CultureTech technology development (R\&D) project for cultural space AX transformation through the Korea Creative Content Agency (KOCCA) grant funded by the Ministry of Culture, Sports and Tourism (MCST) in 2026 (No. RS-2026-25508598).

\clearpage  


%
%

\bibliographystyle{splncs04}
\bibliography{main}



\clearpage
\setcounter{page}{1}
\begin{center}
    \Large
    \textbf{
        \texorpdfstring
        {Dense Reward for Multi-View 3D Reasoning\\with Global Maps and Local Views}
        {Dense Reward for Multi-View 3D Reasoning with Global Maps and Local Views}
    } \\
    \textit{(Supplementary Material)}
\end{center}

\setcounter{tocdepth}{2}                                            
\startcontents[supplement]                                          
\hypersetup{linkcolor=eccvblue}                                     
\printcontents[supplement]{}{1}{\section*{Supplementary Material}}  

\clearpage
\setcounter{section}{0}
\renewcommand{\thesection}{S\arabic{section}}

\setcounter{figure}{0}
\renewcommand{\thefigure}{A\arabic{figure}}

\setcounter{table}{0}
\renewcommand{\thetable}{A\arabic{table}}

\setcounter{equation}{0}
\renewcommand{\theequation}{A\arabic{equation}}

\section{Overview}

This supplementary material provides additional details on the proposed map-grounded learning framework for multi-view 3D reasoning.
In particular, we describe the structured reasoning pipeline, the construction of supervised training data, the implementation and optimization details for SFT and GRPO, and the dataset-specific evaluation protocols used in our experiments.
We also include additional analyses and qualitative examples to complement the results in the main paper.

\section{Limitations \& Future Work}

While this work focuses on static scenes, extending the framework to dynamic environments (e.g., video, 4D) is challenging due to temporal correspondence, motion, and scene changes.
Nevertheless, the proposed formulation can be naturally generalized by defining temporally-aware consistency and exploration rewards.
Another promising direction is multi-turn spatial reasoning.
Iterative querying with memory updates and long-horizon planning could support more precise, compositional 3D inference and improve robustness in complex scenarios.

\section{Method Details}
\label{sec:supp_method}

This section provides additional methodological details deferred from the main paper, including the representation of cognitive maps and the similarity function used for global reward computation.

\begin{figure}[t]
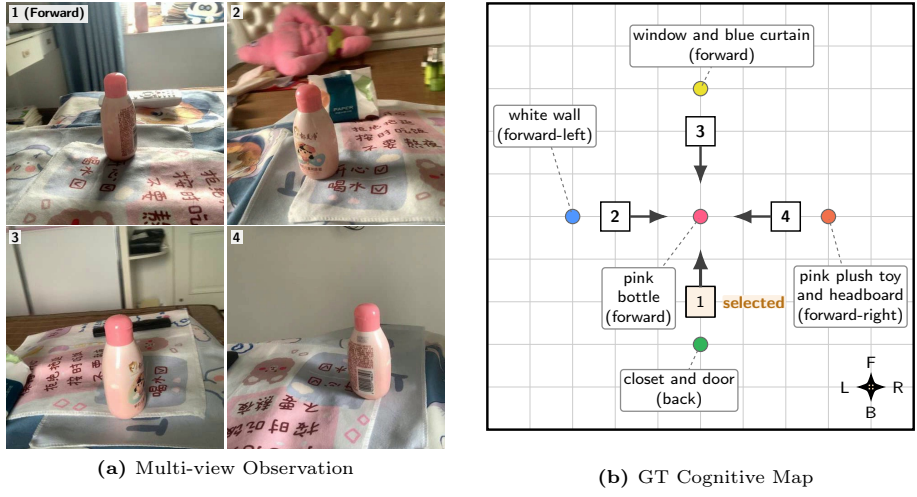

    \centering
    \begin{subfigure}[t]{0.48\textwidth}
        \centering
        \vspace{0pt}
        \input{figures/intro_figure/subfig_1_multiview.tex}
        \caption{Multi-view Observation}
        \label{fig:suppl_left_obervation}
    \end{subfigure}
    \hfill
    \begin{subfigure}[t]{0.48\textwidth}
        \centering
        \vspace{0pt}
        \input{figures/intro_figure/subfig_4_gt_cogmap.tex}
        \caption{GT Cognitive Map}
        \label{fig:suppl_right_cog_map}
    \end{subfigure}
    \caption{
        Example of the allocentric cognitive map representation.
        \textbf{(a)} Multi-view observations of a scene captured from different viewpoints.
        \textbf{(b)} The corresponding ground-truth cognitive map represented on a discretized bird's-eye-view grid, where objects and viewpoints are placed according to their relative spatial arrangement in a world-centered reference frame.
    }
    \label{fig:suppl_main}
\end{figure}

\subsection{Cognitive Map Representation}
\label{subsec:map_representation}

Following previous cognitive-map-based formulations for multi-view spatial reasoning~\cite{wang2026understanding_mindcube, gupta2017_cognitive_map, wang2023_gridmm, yang2025_thinking_in_space_vsi_bench}, we represent the allocentric cognitive map as a structured bird's-eye-view scene abstraction defined on a discretized $10 \times 10$ grid~\cite{yang2025_thinking_in_space_vsi_bench,wang2026understanding_mindcube}.
As shown in~\Cref{fig:suppl_left_obervation,fig:suppl_right_cog_map}, multi-view observations are integrated into a unified allocentric representation, where the resulting cognitive map captures the coarse spatial organization of the scene in a world-centered reference frame.

Each cognitive map consists of two types of entries: \textit{objects}, which denote salient scene entities, and \textit{views}, which denote camera viewpoints associated with the input images.
In practice, each cognitive map is serialized as a lightweight JSON structure containing object and viewpoint entries.
Each object entry contains its semantic name and grid position, and may additionally include a discrete facing attribute when such orientation is available.
Each view entry contains its identifier, grid position, and a discrete facing direction.
We use a small set of canonical orientation labels, namely \emph{up}, \emph{down}, \emph{left}, and \emph{right}, to represent viewpoint orientation on the grid.

This representation is intentionally lightweight and structured.
Rather than modeling metrically precise 3D geometry, it captures the coarse spatial layout that is sufficient for downstream trajectory planning and grounded reasoning.
In particular, the grid coordinates should be interpreted as a compact structural abstraction of the scene, where object and viewpoint locations indicate their relative arrangement in the global map.
The resulting representation is both interpretable and easy for the MLLM to generate as an explicit intermediate state.

Formally, the allocentric cognitive map is represented as:
\begin{equation}
    \mathcal{C}^{\mathrm{allo}} =
    \left\{
    \mathcal{O}^{\mathrm{allo}}, \mathcal{V}^{\mathrm{allo}}
    \right\},
\end{equation}
where $\mathcal{O}^{\mathrm{allo}}$ denotes the set of object entries and $\mathcal{V}^{\mathrm{allo}}$ denotes the set of viewpoint entries.
Each object entry is described by its name and position, while each viewpoint entry is described by its name, position, and facing direction.
This structured map serves as the global intermediate representation on top of which the subsequent local trajectory planning and egocentric grounding are performed.

\subsection{Cognitive Map Similarity for Global Reward}
\label{subsec:pseudo_map_similarity}

Given the predicted allocentric map $\mathcal{C}^{\mathrm{allo}}$ and the pseudo target $\mathcal{C}^{*}$, the global reward is defined by a structural similarity function,
\begin{equation}
    R_{\mathrm{global}} = \texttt{sim}\!\left(\mathcal{C}^{\mathrm{allo}}, \mathcal{C}^{*}\right).
\end{equation}
The similarity score evaluates whether the predicted map preserves the spatial structure of the reference map.
Importantly, the objective does not require exact alignment of absolute coordinates on the discretized $10 \times 10$ grid.
Instead, it emphasizes the \emph{relative} spatial configuration among salient objects in the ground-truth map, since preserving such relational structure is more important for downstream reasoning than reproducing exact positions.

We compute the similarity from two complementary components,
\begin{equation}
    \mathrm{sim}\!\left(\mathcal{C}^{\mathrm{allo}}, \mathcal{C}^{*}\right)
    = \alpha_{\mathrm{sim}} s_{\mathrm{dir}} + (1-\alpha_{\mathrm{sim}}) s_{\mathrm{face}},
\end{equation}
where $s_{\mathrm{dir}} \in [0,1]$ denotes directional similarity, $s_{\mathrm{face}} \in [0,1]$ denotes facing similarity, and $\alpha_{\mathrm{sim}} \in [0,1]$ balances the two terms.

For \textbf{directional similarity}, we first identify the set of salient objects shared by $\mathcal{C}^{\mathrm{allo}}$ and $\mathcal{C}^{*}$, where the reference objects in $\mathcal{C}^{*}$ define the comparison set.
We then compare the coarse pairwise relation of each object pair in the two maps.
Each pair is assigned a discrete relative direction such as \emph{left}, \emph{right}, \emph{up}, \emph{down}, or an inward/outward relation when the pair is spatially coincident under the adopted map convention.
The directional score is given by the fraction of object pairs whose relative relations agree between the two maps.
This formulation makes the reward invariant to global translation and robust to small coordinate perturbations, as long as the relative layout is preserved.

For \textbf{facing similarity}, we compare the facing direction associated with each valid viewpoint in the predicted and reference maps.
The score is defined as the fraction of reference viewpoints whose discrete facing labels match those in the prediction.
Therefore, this term evaluates whether the model not only captures the spatial layout of the scene, but also preserves the viewpoint orientation needed for grounded reasoning.

Overall, the global similarity function favors structurally consistent cognitive maps rather than surface-level coordinate matching.
As a result, the model is rewarded for recovering the relative scene geometry and viewpoint orientation that are most relevant to trajectory planning and 3D reasoning.

\begin{figure}[t]
    \centering

    \begin{subfigure}[b]{0.24\textwidth}
        \centering
        \includegraphics[width=\linewidth]{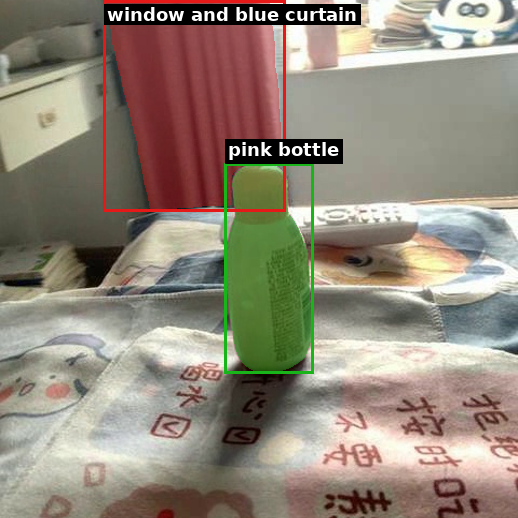}
        \caption{View 1}
    \end{subfigure}
    \hfill
    \begin{subfigure}[b]{0.24\textwidth}
        \centering
        \includegraphics[width=\linewidth]{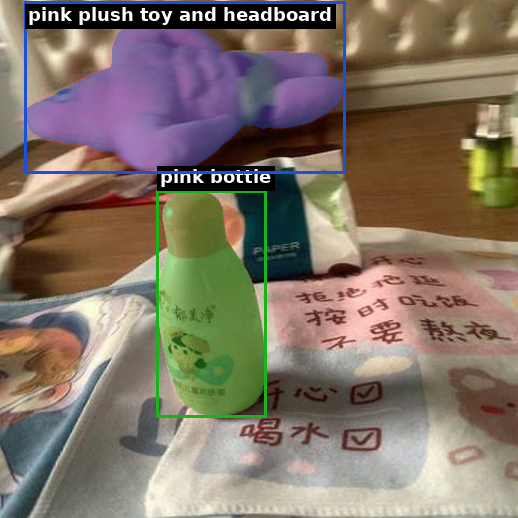}
        \caption{View 2}
    \end{subfigure}
    \hfill
    \begin{subfigure}[b]{0.24\textwidth}
        \centering
        \includegraphics[width=\linewidth]{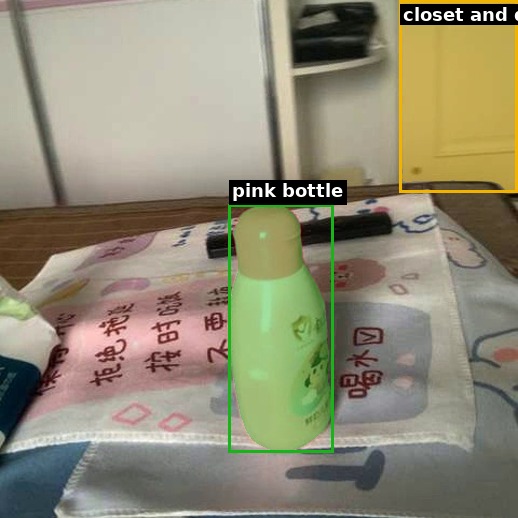}
        \caption{View 3}
    \end{subfigure}
    \hfill
    \begin{subfigure}[b]{0.24\textwidth}
        \centering
        \includegraphics[width=\linewidth]{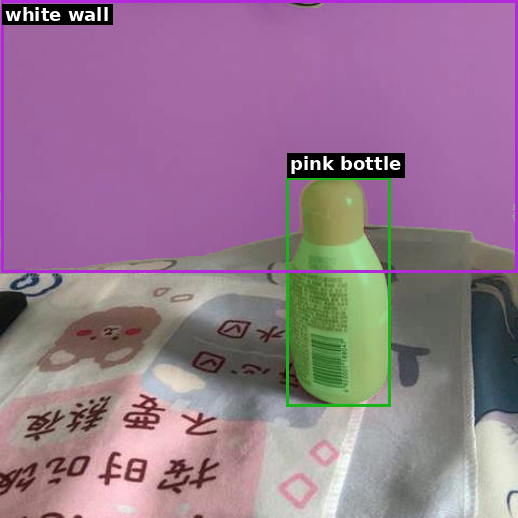}
        \caption{View 4}
    \end{subfigure}

    \begin{subfigure}[b]{0.3\textwidth}
        \centering
        \includegraphics[width=\linewidth]{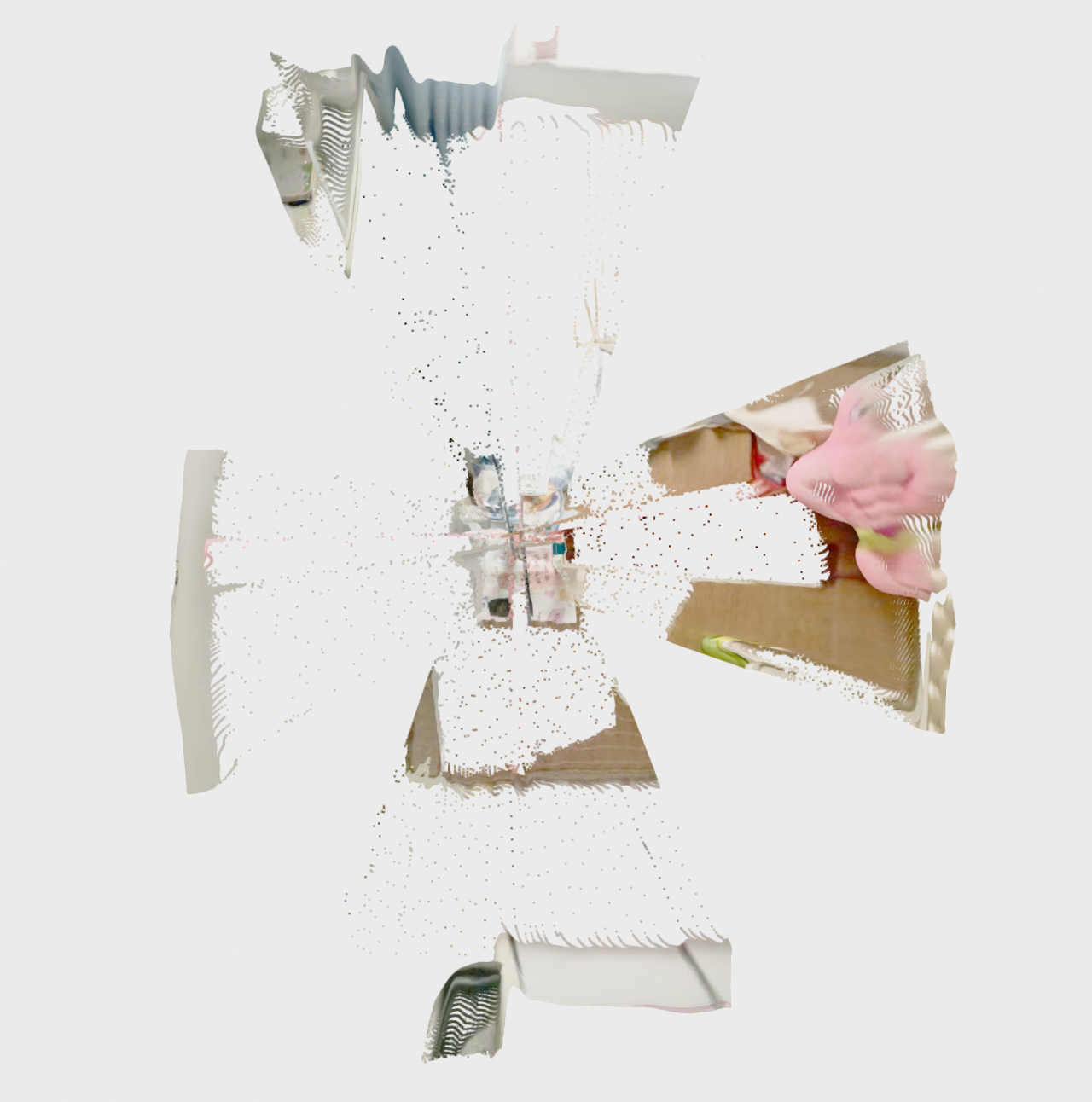}
        \caption{VGGT}
        \label{fig:b}
    \end{subfigure}
    \hfill
    \begin{subfigure}[b]{0.3\textwidth}
        \centering
        \includegraphics[width=\linewidth]{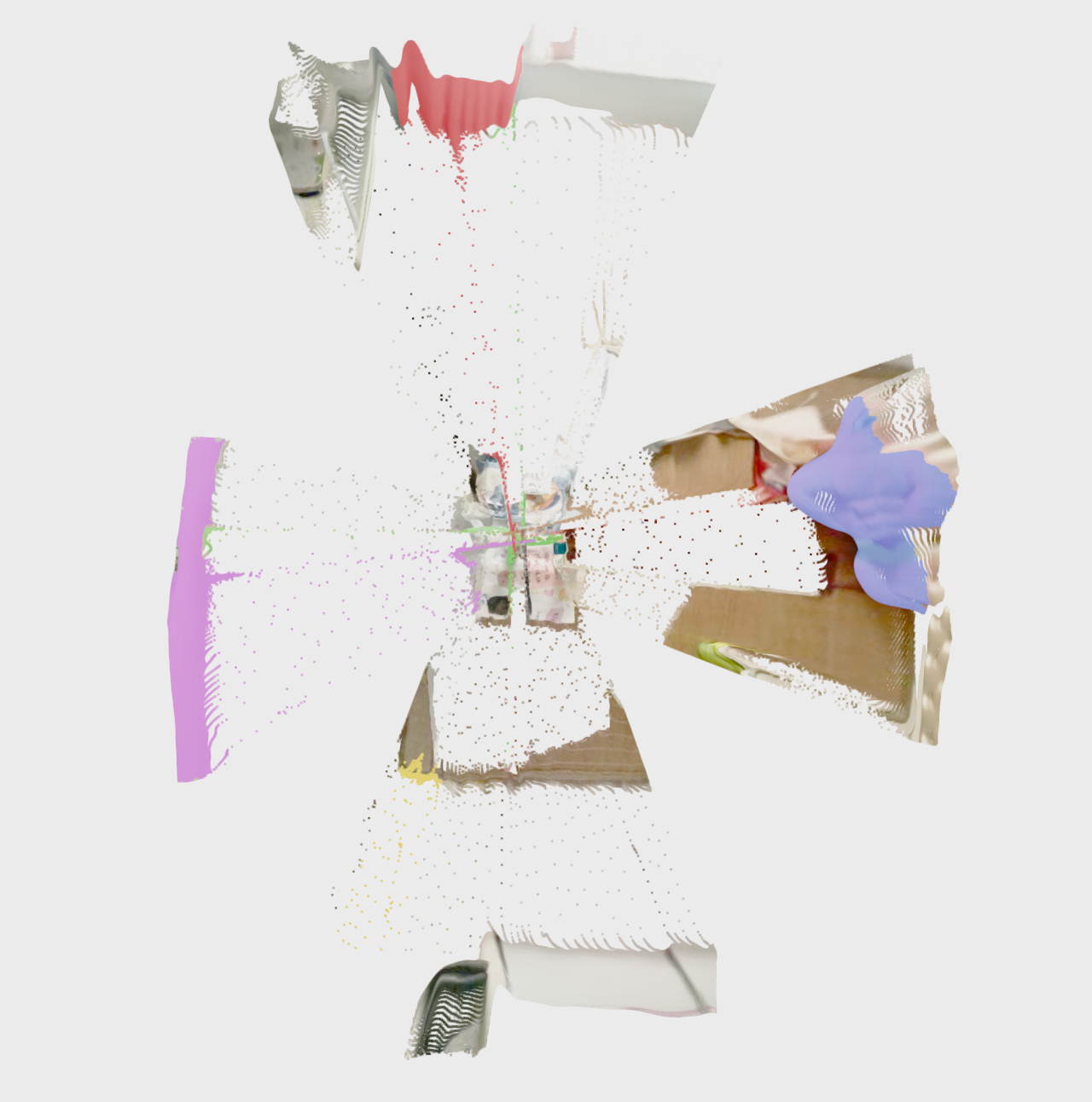}
        \caption{VGGT + SAM3}
        \label{fig:c}
    \end{subfigure}
    \hfill
    \begin{subfigure}[b]{0.3\textwidth}
        \centering
        \includegraphics[width=\linewidth]{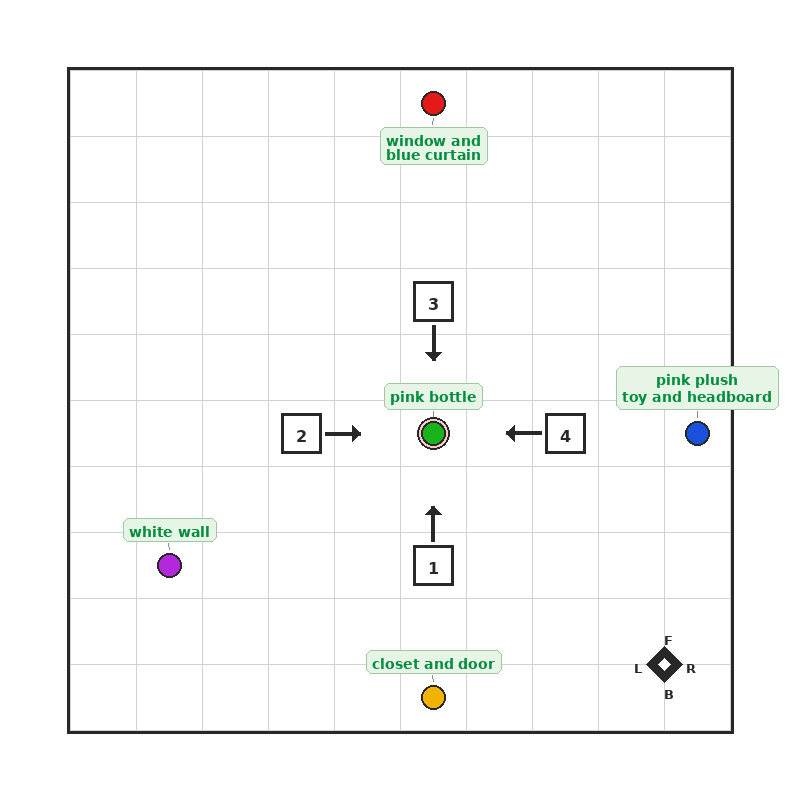}
        \caption{Cognitive Map}
        \label{fig:d}
    \end{subfigure}

    \caption{\textbf{Cognitive map construction from multi-view observations.}
We first segment salient objects in the question with SAM3~\cite{carion2025_sam3} across multi-view images. Then, we reconstruct the scene with 3D vision foundation model such as VGGT~\cite{wang2025_vggt}, enriching it with semantic grounding from 2D segmentation results from multi-views.
The bottom row shows the VGGT reconstruction, the semantically grounded scene after SAM3, and the final cognitive map.
The final cognitive map summarizes the spatial layout in an allocentric bird's-eye view.}
    \label{fig:suppl_cognitive_map}
\end{figure}

\section{Implementation Details}
\label{sec:supp_impl}

Our framework is built upon Qwen2.5-VL~\cite{qwen2025_qwen2_5} and is implemented by extending the official MindCube~\cite{wang2026understanding_mindcube} codebases.
We use Qwen2.5-VL as the main multimodal backbone throughout the training pipeline, allowing us to study the effect of structured supervision and dense reward optimization under a fixed model architecture.
Unless otherwise noted, we use a maximum sequence length of 8,192 tokens so that the model can process multi-view visual inputs, task instructions, intermediate reasoning fields, and the final answer within a unified generation format.
We first train the model with supervised fine-tuning (SFT) on structured targets, and then further optimize it with Group Relative Policy Optimization (GRPO).
For GRPO training, we use the $\texttt{verl}$~\cite{sheng2024hybridflow_verl} framework, which provides an efficient infrastructure for reward-based optimization over large multimodal policies.
In the remainder of this section, we describe the input-output format, the SFT data construction and optimization procedure, the pseudo cognitive map construction, the GRPO setup and reward implementation, and the inference details necessary to reproduce our method.

\subsection{Input-Output Formatting.}
Each training instance consists of a set of multi-view images, a question, and a structured target sequence.
We use a unified autoregressive output format with four blocks:
\[
    \begin{gathered}
        \texttt{<CogMap> ... </CogMap>} \\
        \texttt{<think> ... }\\
        \texttt{<EgoMap> ... </EgoMap>} \\
        \texttt{ ... </think>} \\
        \texttt{<answer> ... </answer>}
    \end{gathered}
\]
This format enables the model to jointly learn intermediate spatial representations and final answer prediction within a single generation stream.
The \texttt{<CogMap>} block represents the global allocentric scene layout, the \texttt{<EgoMap>} block represents the viewpoint-aligned local layout relevant to the question, the \texttt{<think>} block verbalizes the grounded reasoning process, and the \texttt{<answer>} block contains the final answer used for evaluation.
We use a maximum sequence length of 8,192 tokens so that multi-view visual inputs, instructions, intermediate reasoning fields, and the final answer can be processed within a single context window.

\subsection{SFT Data Construction.}

\begin{figure*}[!ht]
  \centering

  \definecolor{PromptHeader}{RGB}{55,55,55}
  \definecolor{ExamplePurple}{RGB}{88,70,150}
  \definecolor{ExampleGreen}{RGB}{0,120,90}

  \begin{minipage}{0.96\linewidth}

    \begin{tcolorbox}[
      colback=PromptHeader,
      colframe=PromptHeader,
      arc=2.2mm,
      boxrule=0pt,
      left=3.2mm,right=3.2mm,top=1.2mm,bottom=6mm,
      before skip=0pt, after skip=0pt
    ]
      {\sffamily\bfseries\fontsize{10}{11}\selectfont\color{white}
      Example of SFT target construction (MindCube)}
    \end{tcolorbox}

    \begin{tcolorbox}[
      colback=white,
      colframe=black!65,
      arc=2.2mm,
      boxrule=0.8pt,
      left=3.2mm,right=3.2mm,top=2.2mm,bottom=2.0mm,
      before skip=-6mm, after skip=0pt,
      before upper=\par,
      fontupper=\ttfamily\scriptsize
    ]


{\color{ExamplePurple}\sffamily\bfseries [Rules]}\\
... \\
{\color{ExamplePurple}\sffamily\bfseries [Question]}\\
... \\

\par\smallskip

{\color{ExampleGreen}\sffamily\bfseries [Answer]}\\
<CogMap>\{\\
\hspace*{1.0em}``objects'': [\\
\hspace*{2.0em}\{``name'': ``black waist bag'', ``position'': [5, 5]\},\\
\hspace*{2.0em}\{``name'': ``plush toy'', ``position'': [5, 8]\},\\
\hspace*{2.0em}\{``name'': ``black sofa'', ``position'': [2, 5], ``facing'': ``down''\},\\
\hspace*{2.0em}\{``name'': ``window'', ``position'': [5, 2]\},\\
\hspace*{2.0em}\{``name'': ``display shelves'', ``position'': [8, 5]\}\\
\hspace*{1.0em}],\\
\hspace*{1.0em}``views'': [\\
\hspace*{2.0em}\{``name'': ``Image 3'', ``position'': [5, 4], ``facing'': ``down''\},\\
\hspace*{2.0em}\{``name'': ``Image 1'', ``position'': [5, 6], ``facing'': ``up''\},\\
\hspace*{2.0em}\{``name'': ``Image 2'', ``position'': [4, 5], ``facing'': ``right''\},\\
\hspace*{2.0em}\{``name'': ``Image 4'', ``position'': [6, 5], ``facing'': ``left''\}\\
\hspace*{1.0em}]\\
\}</CogMap>\\
<think>\\
(1) Selecting a view: The question references Image 3 at [5, 4] facing=``down''. We should consider the viewpoint from Image 3.\\
\\
(2) Egocentric layout:\\
<EgoMap viewpoint=``Image 3''>\\
- black waist bag: ego\_direction=``forward''\\
- plush toy: ego\_direction=``forward''\\
- black sofa: ego\_direction=``forward-right''\\
- window: ego\_direction=``back''\\
- display shelves: ego\_direction=``forward-left''\\
</EgoMap>\\
\\
Answering from the EgoMap: The question asks about ``left''.
In the EgoMap, ``Display shelves'' has \texttt{ego\_direction=``forward-left''}, so it matches ``left''.\\
\\
(3) Confirming in Image 1: Display shelves is visible,\\
consistent with ego\_direction=``forward-left'' in the EgoMap.\\
\\
(4) Therefore, the answer is ``A. Display shelves''.</think>\\
<answer>A. Display shelves.</answer>

    \end{tcolorbox}
  \end{minipage}

  \caption{
    \textbf{Example of a structured SFT target.}
    Starting from benchmark metadata, we programmatically construct a unified supervision target composed of an allocentric cognitive map, egocentric grounding, structured reasoning trace, and final answer.
    The model is trained to autoregressively generate this full trajectory rather than only the final answer.
  }
  \label{fig:sft_example_mindcube}
\end{figure*}

Starting from the original benchmark training examples~\cite{wang2026understanding_mindcube,yang2025_thinking_in_space_vsi_bench,fu2024_blink}, we construct supervised fine-tuning (SFT) targets through a fully programmatic, template-based pipeline following the data construction strategy of MindCube~\cite{wang2026understanding_mindcube}.
Rather than relying on LLM-generated rationales, all intermediate supervision signals are produced deterministically from the provided ground-truth metadata using predefined templates and rule-based transformations.
Each raw sample consists of a multi-view image set, a question, a ground-truth answer, and scene metadata describing object identities, spatial relations, and viewpoint attributes.
Our goal is to convert each sample into a structured reasoning trajectory aligned with the four stages of our framework: allocentric cognitive map construction, egocentric map grounding, grounded reasoning, and final answer prediction.

For each example, we first construct an allocentric cognitive map from the scene metadata and viewpoint annotations.
This representation places all objects and camera viewpoints into a canonical bird's-eye grid in a world-centered coordinate frame, yielding a global scene representation that is consistent across views.
Although the exact placement rules vary slightly across benchmark settings, the construction is deterministic in all cases and explicitly encodes object identities, coarse coordinates, and viewpoint poses.

Next, we derive an egocentric map conditioned on the question.
Specifically, we parse the question to determine the referenced viewpoint as well as any camera transformation implied by the query, such as a rotation or change of facing direction.
Using this parsed view specification, we compute the effective camera pose and convert object locations from the allocentric map into egocentric directions relative to that pose.
The resulting representation specifies which objects lie in directions such as front, back, left, right, or diagonal sectors, and also records the view that most directly supports the answer object.

Based on the allocentric and egocentric representations, we then generate a grounded reasoning chain using a fixed template that mirrors the intended allocentric-to-egocentric reasoning flow.
The reasoning trace first summarizes the global scene layout and camera poses, then identifies the selected viewpoint and any pose update, next describes the egocentric object arrangement, and finally derives the answer through explicit elimination over candidate objects.
To preserve visual grounding, object references in the reasoning trace are annotated with their supporting image view whenever applicable.

Finally, we serialize the constructed supervision into a single structured target sequence consisting of \texttt{<CogMap>}, \texttt{<EgoMap>}, \texttt{<think>}, and \texttt{<answer>}.
These components are concatenated so that the model is trained to autoregressively generate the full intermediate reasoning trajectory, rather than only the final answer.

The complete SFT data pipeline consists of scaffold generation, prompt construction, model-specific format conversion, and dataset patching for training.
Across all stages, the intermediate representations are generated programmatically from benchmark metadata, ensuring consistency, reproducibility, and close alignment with the template-based supervision protocol introduced in MindCube.
Our final SFT corpus contains 10,000 training examples, each associated with 2 to 6 input images.
We also construct a held-out evaluation split from the benchmark tinybench subset containing 1,050 examples.
A representative example of the resulting structured SFT target is shown in~\Cref{fig:sft_example_mindcube}.

\subsection{Pseudo Cognitive Map Construction.}
To provide a geometrically grounded signal for allocentric map generation, we construct a pseudo ground-truth cognitive map $\mathcal{C}^{*}$ from the multi-view image set $\mathcal{I}$ using frozen 3D vision foundation models (VFMs).
When ground-truth allocentric maps are available in the dataset, we use them directly; otherwise, we use the VFM-derived pseudo map as the supervision target.
Concretely, we first apply a pretrained 3D reconstruction model (e.g., VGGT~\cite{wang2025_vggt}) to infer a scene-level geometric representation from the input views, and then use a segmentation model (e.g., SAM3~\cite{carion2025_sam3}) to identify salient object instances and their spatial extents in the reconstructed scene.
The resulting 3D scene is subsequently projected into a structured allocentric representation that summarizes object identities, coarse locations, and spatial relations in a world-centered reference frame.
This process yields a geometry-consistent pseudo target:
\begin{equation}
    \mathcal{C}^{*} = \mathrm{VFMs}(\mathcal{I}) = \mathcal{F}({\mathrm{SAM3}(\mathrm{VGGT}(\mathcal{I}))}),
\end{equation}
where $\mathcal{F}$ denotes a coarse scene-to-map transformation that converts the reconstructed object-centric 3D scene into the allocentric cognitive-map format used by the language model.
In particular, $\mathcal{F}$ abstracts away fine-grained geometry by discretizing object positions into a canonical world-centered layout and summarizing pairwise spatial relations at a coarse semantic level.
The pseudo map is not intended to be a perfect annotation.
Rather, it provides a stable and verifiable proxy supervision signal that encourages the predicted allocentric cognitive map to remain consistent with the underlying 3D scene geometry.
This design is particularly useful because manually annotating global cognitive maps for large-scale multi-view reasoning data is expensive and difficult to standardize.

\subsection{Optimization for Supervised Fine-tuning.}
During supervised fine-tuning (SFT), we optimize the model using the standard next-token cross-entropy objective over the full target sequence.
This trains the model to autoregressively generate the complete reasoning trajectory, including intermediate map tokens and the final answer.
Unless otherwise noted, we use a global batch size of 256, a per-device batch size of 1, a learning rate of \(1\times10^{-5}\), weight decay of 0, and a warmup ratio of 0.03, and train for 3 epochs.
We use AdamW as the optimizer with a cosine learning rate schedule.
Training is conducted on 8 NVIDIA H100 GPUs.

\subsection{GRPO Training and Reward Implementation.}
After supervised fine-tuning, we further optimize the policy using GRPO with the dense trajectory-level reward defined in~\Cref{eq:dense_reward} of the main paper:
\[
    R(\tau)=\lambda_g R_{\mathrm{global}}+\lambda_l R_{\mathrm{local}}+\lambda_a R_{\mathrm{ans}}+\lambda_f R_{\mathrm{fmt}}.
\]
Here, the global reward measures the structural similarity between the predicted allocentric cognitive map and the target cognitive map, while the local reward compares the predicted ordered view trajectory against a reference trajectory and computes the fraction of step-wise matches.
The answer reward is given by exact match on the final answer, and the format reward checks whether the generated output follows the required structured format and contains a valid answer field.
In practice, the format reward helps suppress malformed generations during reinforcement learning, whereas the global and local rewards provide intermediate supervision signals beyond answer-only optimization.

For each input, we sample a group of rollouts from the current policy and compute group-relative standardized advantages following~\Cref{eq:advantages}.
We then optimize the clipped GRPO objective in~\Cref{eq:GRPO_objective}, together with a KL regularization term that constrains the learned policy toward a reference model.
In our implementation, the SFT checkpoint is used both to initialize the policy and as the KL reference policy, and GRPO is implemented using the \texttt{verl} framework.
Unless otherwise noted, we use a rollout group size of \(K=8\), clipping coefficient \(\epsilon=0.2\), KL coefficient \(\beta=0.001\), and a learning rate of \(1\times10^{-6}\), and train for 600 update steps.
The reward weights are set to \(\lambda_g=0.5\), \(\lambda_l=0.2\), \(\lambda_a=0.5\), and \(\lambda_f=0.2\).

\subsection{Inference and Evaluation.}
At inference time, the model receives the same multi-view input format as in training and produces a single structured response.
We extract the final prediction from the \texttt{<answer>} field for benchmark evaluation.
For MindCube and BLINK (MV), we report accuracy.
For VSI-Bench, we report category-wise scores and the overall average following the benchmark protocol.
For VSI-Bench in particular, although the model is trained on static multi-view images, we adapt the inference prompt to sampled video frames by instructing the model to first build an egocentric map from the frames and then follow the same coordinate-based reasoning pipeline used during training.

\subsection{Code Release.}
We will release the training and inference code, together with data preprocessing scripts and evaluation utilities.

\section{Dataset and Evaluation Details}

This section summarizes the benchmark-specific input setting and evaluation protocol used in our experiments.
We evaluate on three benchmarks: MindCube, VSI-Bench, and BLINK (MV).
Following the benchmark conventions, we report accuracy on MindCube and BLINK (MV), and category-wise scores together with the overall average on VSI-Bench.

\begin{figure}[!ht]
  \centering

  \definecolor{PromptHeader}{RGB}{55,55,55}
  \definecolor{PromptGreen}{RGB}{0,120,90}
  \definecolor{PromptBlue}{RGB}{0,92,170}
  \definecolor{PromptOrange}{RGB}{180,110,0}
  \definecolor{PromptRed}{RGB}{210,20,0}

  \begin{minipage}{0.95\linewidth}

    \begin{tcolorbox}[
      colback=PromptHeader,
      colframe=PromptHeader,
      arc=2.2mm,
      boxrule=0pt,
      left=3.2mm,right=3.2mm,top=1.2mm,bottom=6mm,
      before skip=0pt, after skip=0pt
    ]
      {\sffamily\bfseries\fontsize{10}{11}\selectfont\color{white}
      Prompt for MindCube dataset}
    \end{tcolorbox}

    \begin{tcolorbox}[
      colback=white,
      colframe=black!65,
      arc=2.2mm,
      boxrule=0.8pt,
      left=3.2mm,right=3.2mm,top=2.2mm,bottom=2.0mm,
      before skip=-6mm, after skip=0pt,
      before upper=\par,
      fontupper=\ttfamily\scriptsize
    ]

\textcolor{PromptGreen}{\textbf{[System Prompt]}}\\
You are a spatial reasoning assistant for multi-view 3D scenes.
Given multiple images showing a 3D scene from different viewpoints, reason carefully about the spatial layout with constructing an allocentric cognitive map (CogMap) and answer the question.\\
The CogMap is a structured grid where each object has an (x, y) grid coordinate in the allocentric (world-centered) frame, and each viewpoint has a position, facing direction, and associated image.

\par\smallskip

\textcolor{PromptRed}{\textbf{[Answer Instruction]}}\\
Output 3 blocks in order:
\begin{enumerate}\itemsep0.15em
  \item \texttt{<CogMap>}: Construct an allocentric cognitive map from multi-view images.
  \item \texttt{<think>}: Complete a structured reasoning trajectory with 4 numbered steps:
  \begin{enumerate}[label=(\arabic{enumii}), itemsep=0.12em]
    \item Selecting a view: Read the question, identify the spatial anchor, locate it in the CogMap, and determine which viewpoint is most relevant, e.g., \texttt{We should consider the viewpoint from <Image X>.}
    \item Egocentric layout: Convert the allocentric CogMap into an egocentric frame aligned with the selected viewpoint's facing direction. Output as:\\
    \begin{verbatim}
    <EgoMap viewpoint="Image X">
    - [object_name]: ego_direction="[direction]"
    </EgoMap>
    \end{verbatim}
    If rotation/movement is required, apply it and describe updated directions.\\Then, map the question's spatial query to ego-direction labels, e.g., \texttt{A. name=ego\_dir [Image X]}.
    \item Confirmation: Reference another image that visually confirms the answer, e.g., \texttt{Confirming in <Image Y>: [object] is visible, consistent with the EgoMap.}
    \item State the final answer: \texttt{Therefore, the answer is ``X. option\_text''.}
  \end{enumerate}
  \item \texttt{<answer>}: ``X. option\_text''
\end{enumerate}

\par\smallskip

\textcolor{PromptBlue}{\textbf{[CogMap Instruction]}}\\
<cogmap\_gen\_instruction>\\

\par\smallskip

\textcolor{PromptOrange}{\textbf{[egocentric Map Rules]}}\begin{itemize}
  \item The viewpoint's facing direction becomes ``forward'' in the EgoMap.
  \item Objects in the opposite direction become ``back''.
  \item Apply 90$^\circ$ rotation for left/right accordingly.
  \item For diagonal facings, decompose and rotate consistently.
  \item Include only objects that are spatially relevant from the selected viewpoint.
  \item Rotation rules: Grid [0,0]=top-left, [9,9]=bottom-right.
  \begin{itemize}\itemsep0.08em
    \item facing=up $\rightarrow$ forward=[-y], right=[+x]
    \item facing=right $\rightarrow$ forward=[+x], right=[+y]
    \item facing=down $\rightarrow$ forward=[+y], right=[-x]
    \item facing=left $\rightarrow$ forward=[-x], right=[-y]
  \end{itemize}
\end{itemize}
    \end{tcolorbox}
  \end{minipage}

  \caption{
    \textbf{Prompt template for MindCube-Tiny.}
    We follow the prompts of MindCube for allocentric cognitive map generation, and add explicit egocentric representations.
    }
  \label{fig:template_mindcube}
\end{figure}

\subsection{MindCube}
MindCube evaluates spatial reasoning under partial observations and dynamic viewpoints using multi-view image groups paired with multiple-choice spatial questions.
The benchmark is organized around three camera movement settings: \textit{Rotation}, \textit{Among}, and \textit{Around}, which respectively test in-place view rotation, movement among objects, and movement around the scene.
The questions cover viewpoint-dependent spatial relations, perspective taking, and hypothetical motion, and are designed to require reasoning about objects that may not be visible in the current view.
In our experiments, we follow the MindCube-Tiny protocol used in the main paper, which contains 10K training examples and a 1K evaluation split; for brevity, we denote this evaluation set as \emph{MindCube} throughout the paper.
We use the benchmark-specific prompting template shown in~\Cref{fig:template_mindcube}, and report accuracy on each of the three settings and the overall average.

\begin{figure}[!ht]
  \centering

  \definecolor{PromptHeader}{RGB}{55,55,55}
  \definecolor{PromptGreen}{RGB}{0,120,90}
  \definecolor{PromptBlue}{RGB}{0,92,170}
  \definecolor{PromptOrange}{RGB}{180,110,0}
  \definecolor{PromptRed}{RGB}{210,20,0}

  \begin{minipage}{0.95\linewidth}

    \begin{tcolorbox}[
      colback=PromptHeader,
      colframe=PromptHeader,
      arc=2.2mm,
      boxrule=0pt,
      left=3.2mm,right=3.2mm,top=1.2mm,bottom=6mm,
      before skip=0pt, after skip=0pt
    ]
      {\sffamily\bfseries\fontsize{10}{11}\selectfont\color{white}
      Prompt for VSI-Bench evaluation}
    \end{tcolorbox}

    \begin{tcolorbox}[
      colback=white,
      colframe=black!65,
      arc=2.2mm,
      boxrule=0.8pt,
      left=3.2mm,right=3.2mm,top=2.2mm,bottom=2.0mm,
      before skip=-6mm, after skip=0pt,
      before upper=\par,
      fontupper=\ttfamily\scriptsize
    ]

\textcolor{PromptGreen}{\textbf{[System Prompt]}}\\
You are a spatial reasoning assistant for 3D scenes captured as a multi-frame video.
Given frames uniformly sampled from a video of a 3D scene, construct an allocentric cognitive map (CogMap) from the visual observations and reason about the spatial layout to answer the question.\\

\par\smallskip

\textcolor{PromptRed}{\textbf{[Answer Instruction]}}\\
Output 3 blocks in order:
\begin{enumerate}\itemsep0.15em
  \item \texttt{<CogMap>}: Construct an allocentric cognitive map as a 10$\times$10 grid by observing the video frames. Assign (x, y) coordinates to each object based on their estimated positions in the scene.
  \item \texttt{<think>}: Complete a structured reasoning trajectory with 4 numbered steps:
  \begin{enumerate}[label=(\arabic{enumii}), itemsep=0.12em]
    \item Selecting a viewpoint: Identify the standing position and facing target in the CogMap from the question, and compute the facing direction.
    \item Egocentric layout: Convert the allocentric CogMap into an egocentric frame aligned with the standing position's facing direction. Output as:\\
    \begin{verbatim}
    <EgoMap position="[object_name]">
    - [object_name]: ego_direction="[direction]"
    </EgoMap>
    \end{verbatim}
    From the standing position, compute each direction to relevant objects.
    \item Confirmation: Reference the video frames to visually confirm the spatial relationship.
    \item State the final answer: \texttt{Therefore, the answer is ``X. option\_text''.}
  \end{enumerate}
  \item \texttt{<answer>}:  ``X. option\_text''.
\end{enumerate}

\par\smallskip

\textcolor{PromptBlue}{\textbf{[CogMap Instruction]}}\\
<cogmap\_gen\_instruction>\\

\par\smallskip

\textcolor{PromptOrange}{\textbf{[Egocentric Map Rules]}}\begin{itemize}
  \item The facing direction (from standing position toward the target object) becomes ``forward'' in the ego frame.
  \item Compute displacement vectors from the standing position to each queried object.
  \item Rotation rules: Grid [0,0]=top-left, [9,9]=bottom-right.
  \begin{itemize}\itemsep0.08em
    \item facing=up $\rightarrow$ forward=[-y], right=[+x]
    \item facing=right $\rightarrow$ forward=[+x], right=[+y]
    \item facing=down $\rightarrow$ forward=[+y], right=[-x]
    \item facing=left $\rightarrow$ forward=[-x], right=[-y]
  \end{itemize}
\end{itemize}
    \end{tcolorbox}
  \end{minipage}

  \caption{
    \textbf{Prompt template for VSI-Bench evaluation.} Since VSI-Bench provides multi-frame video inputs with object-centric, position-based questions with objects rather than explicit camera viewpoints, we adapt the MindCube-Tiny prompt accordingly.
    }
  \label{fig:template_vsibench}
\end{figure}

\subsection{VSI-Bench}
\label{subsec:details_vsi_bench}

VSI-Bench is a video-based visual-spatial intelligence benchmark built from real indoor-scene videos with object-level 3D annotations, covering a diverse set of spatial reasoning tasks such as object count, relative distance, relative direction, route planning, object size, absolute distance, room size, and appearance order.
Unlike MindCube, VSI-Bench poses questions grounded in physical positions and orientations in the scene (e.g., ``standing by the door and facing the bookshelf'') rather than referencing specific camera views.
In addition, the visual input is given as uniformly sampled video frames instead of discrete multi-view images with known viewpoints.
To bridge this domain gap, we adapt the inference prompt to instruct the model to first construct an egocentric map from the sampled frames, and then preserve the same coordinate-based egocentric reasoning pipeline used during training on MindCube-Tiny, as illustrated in~\Cref{fig:template_vsibench}.
This adaptation is important because VSI-Bench differs from our training setup in both the question format and the visual input distribution.
Following the benchmark protocol, we report category-wise results together with the overall average score.

\begin{figure}[!ht]
  \centering

  \definecolor{PromptHeader}{RGB}{55,55,55}
  \definecolor{PromptGreen}{RGB}{0,120,90}
  \definecolor{PromptBlue}{RGB}{0,92,170}
  \definecolor{PromptOrange}{RGB}{180,110,0}
  \definecolor{PromptRed}{RGB}{210,20,0}

  \begin{minipage}{0.95\linewidth}

    \begin{tcolorbox}[
      colback=PromptHeader,
      colframe=PromptHeader,
      arc=2.2mm,
      boxrule=0pt,
      left=3.2mm,right=3.2mm,top=1.2mm,bottom=6mm,
      before skip=0pt, after skip=0pt
    ]
      {\sffamily\bfseries\fontsize{10}{11}\selectfont\color{white}
      Prompt for BLINK (MV) evaluation}
    \end{tcolorbox}

    \begin{tcolorbox}[
      colback=white,
      colframe=black!65,
      arc=2.2mm,
      boxrule=0.8pt,
      left=3.2mm,right=3.2mm,top=2.2mm,bottom=2.0mm,
      before skip=-6mm, after skip=0pt,
      before upper=\par,
      fontupper=\ttfamily\scriptsize
    ]

\textcolor{PromptGreen}{\textbf{[System Prompt]}}\\
You are a spatial reasoning assistant for multi-view 3D scenes.
Given multiple images showing a 3D scene from different viewpoints, reason carefully about the spatial layout with constructing an allocentric cognitive map (CogMap) and answer the question.\\
The CogMap is a structured grid where each object has an (x, y) grid coordinate in the allocentric (world-centered) frame, and each viewpoint has a position, facing direction, and associated image.

\par\smallskip

\textcolor{PromptRed}{\textbf{[Answer Instruction]}}\\
Output 3 blocks in order:
\begin{enumerate}\itemsep0.15em
  \item \texttt{<CogMap>}: Construct an allocentric cognitive map from multi-view images.
  \item \texttt{<think>}: Complete a structured reasoning trajectory with 4 numbered steps:
  \begin{enumerate}[label=(\arabic{enumii}), itemsep=0.12em]
    \item Selecting a view: Read the question, identify the spatial anchor, locate it in the CogMap, and determine which viewpoint is most relevant, e.g., \texttt{We should consider the viewpoint from <Image X>.}
    \item Egocentric layout: Convert the allocentric CogMap into an egocentric frame aligned with the selected viewpoint's facing direction. Output as:\\
    \begin{verbatim}
    <EgoMap viewpoint="Image X">
    - [object_name]: ego_direction="[direction]"
    </EgoMap>
    \end{verbatim}
    If rotation/movement is required, apply it and describe updated directions.
    For camera movement, compare the viewpoint positions across images in the CogMap to determine the direction of movement.\\
    Then, map the question's spatial query to ego-direction labels, e.g., \texttt{A. name=ego\_dir [Image X]}.
    \item Confirmation: Reference another image that visually confirms the answer, e.g., \texttt{Confirming in <Image Y>: [object] is visible, consistent with the EgoMap.}
    \item State the final answer: \texttt{Therefore, the answer is ``X. option\_text''.}
  \end{enumerate}
  \item \texttt{<answer>}: ``X. option\_text''
\end{enumerate}

\par\smallskip

\textcolor{PromptBlue}{\textbf{[CogMap Instruction]}}\\
<cogmap\_gen\_instruction>\\

\par\smallskip

\textcolor{PromptOrange}{\textbf{[Egocentric Map Rules]}}
\begin{itemize}
  \item The viewpoint's facing direction becomes ``forward'' in the EgoMap.
  \item Objects in the opposite direction become ``back''.
  \item Apply 90$^\circ$ rotation for left/right accordingly.
  \item For diagonal facings, decompose and rotate consistently.
  \item Include only objects that are spatially relevant from the selected viewpoint.
  \item Rotation rules: Grid [0,0]=top-left, [9,9]=bottom-right.
  \begin{itemize}\itemsep0.08em
    \item facing=up $\rightarrow$ forward=[-y], right=[+x]
    \item facing=right $\rightarrow$ forward=[+x], right=[+y]
    \item facing=down $\rightarrow$ forward=[+y], right=[-x]
    \item facing=left $\rightarrow$ forward=[-x], right=[-y]
  \end{itemize}
\end{itemize}
    \end{tcolorbox}
  \end{minipage}

  \caption{
  \textbf{Prompt template for BLINK (MV) evaluation.} We adapt a similar MindCube-Tiny prompt as shown in~\Cref{fig:template_mindcube} with additional guidance for camera movement questions, where the model compares viewpoint positions across images in the CogMap.
    }
  \label{fig:template_blink}
\end{figure}

\subsection{BLINK-MV}

BLINK is a multiple-choice benchmark for visual perception that contains 14 tasks spanning low-level correspondence, geometric perception, and higher-level visual understanding.
In this work, we use the multi-view reasoning subset, denoted as BLINK (MV), which evaluates whether a model can infer camera or scene changes from multiple images under partial and distributed visual evidence.
As in the original benchmark, answer choices are discrete options and evaluation is based on exact-match accuracy after mapping the generated response to one of the candidate choices.
We use the prompting template shown in~\Cref{fig:template_blink} and follow the standard BLINK evaluation protocol, reporting accuracy on this multi-view subset.

\section{Additional Experimental Results}

\newcommand{\mvimgsmall}[3]{%
    \begin{tikzpicture}
        \node[inner sep=0, outer sep=0] (img)
        {\includegraphics[width=#3,clip,trim={0 0 0 0}]{#1}};
        \path[use as bounding box] (img.south west) rectangle (img.north east);
        \node[
            anchor=north west,
            fill=white,
            fill opacity=0.88,
            text opacity=1,
            inner sep=0.5pt,
            font=\fontsize{8}{9}\selectfont\sffamily\bfseries
        ] at ([xshift=1pt,yshift=-1pt]img.north west) {#2};
    \end{tikzpicture}%
}

\begin{figure}[!ht]
    \centering

    \newlength{\qualCapHA}
    \setlength{\qualTopA}{0.33\linewidth}
    \setlength{\qualTopB}{0.26\linewidth}
    \setlength{\qualTopC}{0.35\linewidth}
    \setlength{\qualTopH}{0.27\textwidth}  
    \setlength{\qualCapH}{1.8em}
    \setlength{\qualImgH}{\dimexpr\qualTopH-\qualCapH\relax}

    \newlength{\qualTopHA}
    \newlength{\qualImgHA}
    \setlength{\qualCapHA}{2.9em}
    \setlength{\qualTopHA}{0.30\textwidth}
    \setlength{\qualImgHA}{\dimexpr\qualTopHA-\qualCapHA\relax}

    \begin{tcolorbox}[
        width=\linewidth,
        colback=black!1,
        colframe=black!70,
        boxrule=0.6pt,
        arc=1.2pt,
        before skip=0pt,
        after skip=3pt,
        left=1.5mm,right=1.5mm,top=1.1mm,bottom=1.1mm,
        title=\textcolor{white}{\texttt{\textbf{\scriptsize among\_32c0925b\_gen\_6\_1 (MindCube)}}},
        fonttitle=\sffamily\bfseries,
        coltitle=black
    ]

        \makebox[\linewidth][c]{%
            \begin{minipage}[t][\qualTopHA][t]{\qualTopA}
            \centering
        
            \begin{minipage}[c][\qualImgHA][c]{\linewidth}
                \centering
                \setlength{\tabcolsep}{0pt}
                \renewcommand{\arraystretch}{0}
                \adjustbox{max width=0.98\linewidth, max totalheight=\qualImgH, keepaspectratio}{%
                    \begin{tabular}{@{}cc@{}}
                            \mvimgsmall{}{1}{0.47\linewidth} &
                            \mvimgsmall{}{2}{0.47\linewidth} \\
                            \mvimgsmall{}{3}{0.47\linewidth} &
                            \mvimgsmall{}{4}{0.47\linewidth}
                    \end{tabular}
                }
            \end{minipage}
        
            \begin{minipage}[b][\qualCapH][c]{\linewidth}
                \centering
                {\tiny (a) Multi-view images}
            \end{minipage}
        \end{minipage}\hfill

            \begin{minipage}[t][\qualTopH][t]{\qualTopB}
                \centering
                \begin{minipage}[c][\qualImgH][c]{\linewidth}
                    \centering
                    \adjustbox{max width=0.98\linewidth, max totalheight=\qualImgH, keepaspectratio}{%
                        \includegraphics[trim={6 6 6 6},clip]{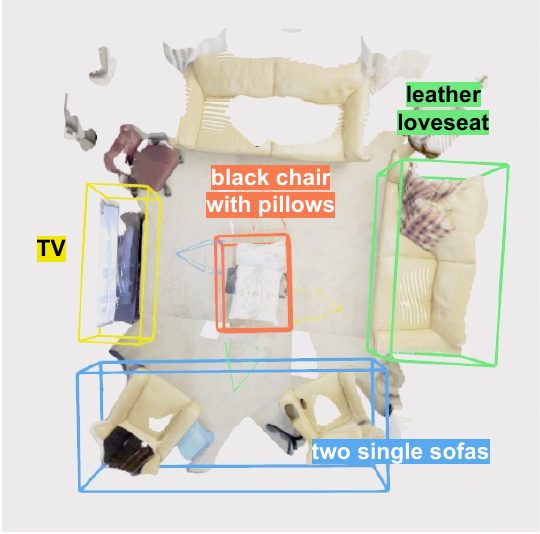}%
                    }
                \end{minipage}

                \begin{minipage}[b][\qualCapH][c]{\linewidth}
                    \centering
                    {\tiny (b) Scene from VFM}
                \end{minipage}
            \end{minipage}\hfill
            \begin{minipage}[t][\qualTopH][t]{\qualTopC}
                \centering
                \begin{minipage}[c][\qualImgH][c]{\linewidth}
                    \centering
                    \adjustbox{max width=0.97\linewidth, max totalheight=\qualImgH, keepaspectratio}{%
\begin{tikzpicture}[font=\fontsize{5.4}{6.0}\selectfont\sffamily]
  \begin{scope}[x=0.46cm,y=-0.46cm]
    \foreach \x in {0,...,10} {\draw[gray!40,line width=0.28pt] (\x,0) -- (\x,10);}
    \foreach \y in {0,...,10} {\draw[gray!40,line width=0.28pt] (0,\y) -- (10,\y);}
    \draw[black,line width=0.75pt] (0,0) rectangle (10,10);

    \definecolor{selView}{RGB}{226,140,28}
    \definecolor{selCorrect}{RGB}{60,154,86}

    \tikzset{
      objdot/.style={draw=black!60,line width=0.35pt,circle,minimum size=4.4pt,inner sep=0pt},
      objlabel/.style={draw=black!45,fill=white,rounded corners=1.0pt,inner sep=1.0pt,align=center},
      callout/.style={black!55,line width=0.34pt,dash pattern=on 1pt off 1pt},
      viewone/.style={draw=black, fill=white, line width=0.45pt},
      viewselect/.style={draw=black, fill=selView!12, line width=0.45pt},
      viewnum/.style={font=\bfseries\fontsize{5.0}{5.0}\selectfont\sffamily}
    }

    \definecolor{colChair}{RGB}{241, 113, 75}
    \definecolor{colSofas}{RGB}{80,150,255}
    \definecolor{colTV}{RGB}{235,225,48}
    \definecolor{colLoveseat}{RGB}{50,180,90}

    \node[objdot,fill=colChair]          (objChair)         at (5,5) {};
    \node[objdot,fill=colSofas]          (objSofas)         at (5,8) {};
    \node[objdot,fill=colTV]             (objTV)            at (2,5) {};

    \node[objdot,fill=colLoveseat]       (objLoveseat)      at (8,4) {};

    \coordinate (labChairPos)      at (2.40,6.5);
    \coordinate (labSofasPos)      at (5,9.10);
    \coordinate (labTVPos)         at (1.8,3.5);
    \coordinate (labLoveseatPos)   at (8.2,2.5);

    \draw[callout,shorten <=0.6pt,shorten >=0.8pt] (labChairPos)     -- (objChair.center);
    \draw[callout,shorten <=0.6pt,shorten >=0.8pt] (labSofasPos)     -- (objSofas.center);
    \draw[callout,shorten <=0.6pt,shorten >=0.8pt] (labTVPos)        -- (objTV.center);
    \draw[callout,shorten <=0.6pt,shorten >=0.8pt] (labLoveseatPos)  -- (objLoveseat.center);

    \node[objlabel] at (labChairPos)
      {black chair\\with pillows\\(forward)};

    \node[objlabel] at (labSofasPos)
      {two single sofas\\(back)};

    \node[objlabel] at (labTVPos)
      {TV\\(forward-left)};

    \node[objlabel] at (labLoveseatPos)
      {leather loveseat\\(forward-right)};

    \def\boxSq{0.34}

    \path[viewselect] (5-\boxSq,7-\boxSq) rectangle (5+\boxSq,7+\boxSq);
    \node[viewnum] at (5,7) {1};
    \node[fill=selView!12,text=selView!80!black,rounded corners=0.6pt,inner sep=0.25pt,anchor=west]
      at (5.5,7) {\textbf{selected}};
    \draw[-{Latex[length=2.1mm]},black!75,line width=0.75pt]
      (5,7-\boxSq) -- (5,5.75);

    \path[viewone] (3-\boxSq,5-\boxSq) rectangle (3+\boxSq,5+\boxSq);
    \node[viewnum] at (3,5) {2};
    \draw[-{Latex[length=2.05mm]},black!75,line width=0.75pt]
      (3+\boxSq,5) -- (4.25,5);

    \path[viewone] (5-\boxSq,3-\boxSq) rectangle (5+\boxSq,3+\boxSq);
    \node[viewnum] at (5,3) {3};
    \draw[-{Latex[length=2.05mm]},black!75,line width=0.75pt]
      (5,3+\boxSq) -- (5,4.25);

    \path[viewone] (7-\boxSq,5-\boxSq) rectangle (7+\boxSq,5+\boxSq);
    \node[viewnum] at (7,5) {4};
    \draw[-{Latex[length=2.05mm]},black!75,line width=0.75pt]
      (7-\boxSq,5) -- (5.75,5);

    \coordinate (egoC) at (9.00,9.00);
    \fill[selView] (egoC) circle (0.05);
    \draw[-{Latex[length=1.4mm]},line width=0.55pt] (egoC) -- ++(0.36,0);   
    \draw[-{Latex[length=1.4mm]},line width=0.55pt] (egoC) -- ++(-0.36,0);  
    \draw[-{Latex[length=1.4mm]},line width=0.55pt] (egoC) -- ++(0,-0.36);  
    \draw[-{Latex[length=1.4mm]},line width=0.55pt] (egoC) -- ++(0,0.36);   
    \node[anchor=west]  at ($(egoC)+(0.25,0)$) {R};
    \node[anchor=east]  at ($(egoC)+(-0.25,0)$) {L};
    \node[anchor=south] at ($(egoC)+(0,-0.2)$) {F};
    \node[anchor=north] at ($(egoC)+(0,0.2)$) {B};
  \end{scope}
\end{tikzpicture}%
                    }
                \end{minipage}

                \begin{minipage}[b][\qualCapH][c]{\linewidth}
                    \centering
                    {\tiny (c) CogMap with ego-directions}
                \end{minipage}
            \end{minipage}%
        }

        \par\smallskip

        \begin{tcolorbox}[
            width=\linewidth,
            colback=black!2,
            colframe=black!55,
            boxrule=0.45pt,
            arc=0.9pt,
            before skip=1pt,
            after skip=5pt,
            left=1.1mm,right=1.1mm,top=0.5mm,bottom=0.5mm,
            fontupper=\fontsize{6}{7}\selectfont\sffamily
        ]
        \textbf{Question:} If I was positioned where the two single sofas is and facing the same direction, what would be to the left of the black chair with pillows from this view?\\
        \textbf{Choices:} A. Leather loveseat with three seat cushions \quad B. Leather loveseat \quad \textbf{C. TV (GT)}
        \end{tcolorbox}

        \begin{tcolorbox}[
            width=\linewidth,
            colback=red!3,
            colframe=red!70!black,
            boxrule=0.45pt,
            arc=0.9pt,
            left=1.1mm,right=1.1mm,top=0.5mm,bottom=0.5mm,
            fontupper=\fontsize{6}{7}\selectfont\sffamily
        ]
        \textbf{Qwen2.5-VL-3B-Instruct:} \texttt{<think>} Let's analyze the information provided:\\
        
        1. The black chair with pillows is placed in front of a television.\\
        2. There are two yellow leather sofas in the room.\\
        3. The black chair is positioned between the two sofas.\\

        From the perspective of someone standing at the location of the two single sofas and facing the same direction:\\
        - The black chair with pillows would be directly in front of the person.\\
        - To the left of the black chair, there would be one of the yellow 
          leather sofas.\\
        
        Therefore, if you were positioned where the two single sofas is and facing the same direction, the object to the left of the black chair with pillows would be the yellow leather sofa.\\
        
        \textbf{Answer}: \textbf{\textcolor{red!75!black}{B}} \\
        \texttt{</think>}\\

        \texttt{<answer>}\textbf{\textcolor{red!75!black}{B. Leather loveseat}}\texttt{</answer>}
        \end{tcolorbox}

        \begin{tcolorbox}[
            width=\linewidth,
            colback=green!4,
            colframe=green!45!black,
            boxrule=0.45pt,
            arc=0.9pt,
            left=1.1mm,right=1.1mm,top=0.5mm,bottom=0.5mm,
            fontupper=\fontsize{6}{7}\selectfont\sffamily
        ]
        \textbf{Ours:} \texttt{<think>}\\ 
        
        \textbf{(1) Selecting a view:}\\
        Two single sofas are facing the same direction as in \textbf{\underline{Image 1}}, so we should consider the viewpoint from \textbf{\underline{Image 1}}.\\
        
        \textbf{(2) Egocentric layout:}\\
        The question asks about ``left''. In the EgoMap, ``TV'' has \texttt{ego\_direction="forward-left"}, so it matches ``left''. \\
        
        \textbf{(3) Confirming in \textbf{\underline{Image 4:}}}\\
        TV can be checked in Image 4 which direction is also facing left in the EgoMap.\\
        
        \textbf{(4)} Therefore, the answer is \textbf{\textcolor{green!25!black}{``C. TV''}}. \\
        \texttt{</think>}\\

        \texttt{<answer>}\textbf{\textcolor{green!25!black}{C. TV}}\texttt{</answer>}
        \end{tcolorbox}

    \end{tcolorbox}
    \caption{
        \textbf{Qualitative comparison on MindCube-Tiny (Among).}
    }
    \label{fig:quali_mindcube_among_01}
\end{figure}

\begin{figure}[!ht]
    \centering

    \setlength{\qualTopA}{0.33\linewidth}
    \setlength{\qualTopB}{0.26\linewidth}
    \setlength{\qualTopC}{0.35\linewidth}
    \setlength{\qualTopH}{0.27\textwidth}  
    \setlength{\qualCapH}{1.8em}
    \setlength{\qualImgH}{\dimexpr\qualTopH-\qualCapH\relax}

    \setlength{\qualCapHA}{2.9em}
    \setlength{\qualTopHA}{0.30\textwidth}
    \setlength{\qualImgHA}{\dimexpr\qualTopHA-\qualCapHA\relax}

    \begin{tcolorbox}[
        width=\linewidth,
        colback=black!1,
        colframe=black!70,
        boxrule=0.6pt,
        arc=1.2pt,
        before skip=0pt,
        after skip=3pt,
        left=1.5mm,right=1.5mm,top=1.1mm,bottom=1.1mm,
        title=\textcolor{white}{\texttt{\textbf{\scriptsize rotation\_group015\_q2\_3 (MindCube)}}},
        fonttitle=\sffamily\bfseries,
        coltitle=black
    ]

        \makebox[\linewidth][c]{%
            \begin{minipage}[t][\qualTopHA][t]{\qualTopA}
            \centering
        
            \begin{minipage}[c][\qualImgHA][c]{\linewidth}
                \centering
                \setlength{\tabcolsep}{0pt}
                \renewcommand{\arraystretch}{0}
                \adjustbox{max width=1.0\linewidth, max totalheight=\qualImgH, keepaspectratio}{%
                    \begin{tabular}{@{}cc@{}}
                             \mvimgsmall{}{1}{0.32\linewidth} &
                            \mvimgsmall{}{2}{0.32\linewidth} \\
                            \mvimgsmall{}{3}{0.32\linewidth} 
                    \end{tabular}
                }
            \end{minipage}
        
            \begin{minipage}[b][\qualCapH][c]{\linewidth}
                \centering
                {\tiny (a) Multi-view images}
            \end{minipage}
        \end{minipage}\hfill

            \begin{minipage}[t][\qualTopH][t]{\qualTopB}
                \centering
                \begin{minipage}[c][\qualImgH][c]{\linewidth}
                    \centering
                    \adjustbox{max width=0.98\linewidth, max totalheight=\qualImgH, keepaspectratio}{%
                        \includegraphics[trim={6 6 6 6},clip]{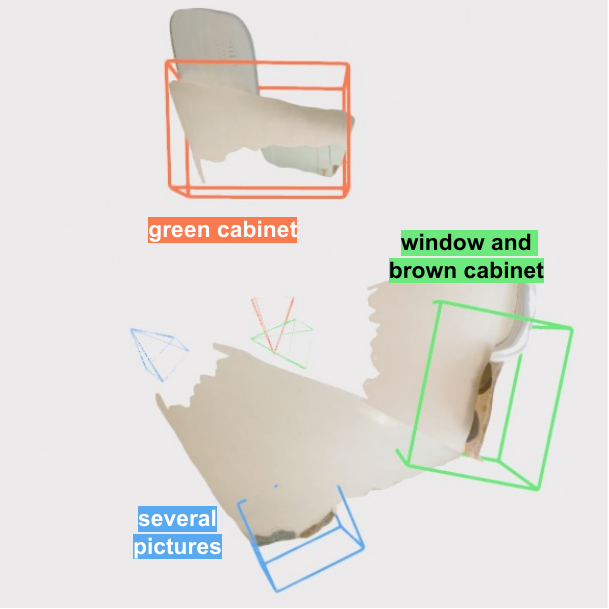}%
                    }
                \end{minipage}

                \begin{minipage}[b][\qualCapH][c]{\linewidth}
                    \centering
                    {\tiny (b) Scene from VFM}
                \end{minipage}
            \end{minipage}\hfill
            \begin{minipage}[t][\qualTopH][t]{\qualTopC}
                \centering
                \begin{minipage}[c][\qualImgH][c]{\linewidth}
                    \centering
                    \adjustbox{max width=0.97\linewidth, max totalheight=\qualImgH, keepaspectratio}{%
\begin{tikzpicture}[font=\fontsize{5.4}{6.0}\selectfont\sffamily]
  \begin{scope}[x=0.46cm,y=-0.46cm]
    \foreach \x in {0,...,10} {\draw[gray!40,line width=0.28pt] (\x,0) -- (\x,10);}
    \foreach \y in {0,...,10} {\draw[gray!40,line width=0.28pt] (0,\y) -- (10,\y);}
    \draw[black,line width=0.75pt] (0,0) rectangle (10,10);

    \definecolor{selView}{RGB}{226,140,28}

    \tikzset{
      objdot/.style={draw=black!60,line width=0.35pt,circle,minimum size=4.4pt,inner sep=0pt},
      objlabel/.style={draw=black!45,fill=white,rounded corners=1.0pt,inner sep=1.0pt,align=center},
      callout/.style={black!55,line width=0.34pt,dash pattern=on 1pt off 1pt},
      innerwhite/.style={fill=white, draw=black, line width=0.25pt},
      viewone/.style={draw=black, fill=white, line width=0.45pt},
      viewnum/.style={font=\bfseries\fontsize{5.0}{5.0}\selectfont\sffamily}
    }

    \definecolor{colCabinet}{RGB}{241, 113, 75}       
    \definecolor{colPic}{RGB}{80,150,255}    
    \definecolor{colWindow}{RGB}{50, 180, 90}       

    \node[objdot,fill=colCabinet]    (objCabinet)  at (5,2) {};
    \node[objdot,fill=colPic] (colPic)  at (4,8) {};
    \node[objdot,fill=colWindow]   (objWin) at (8,5) {};

    
    \coordinate (labCabinetPos)  at (2.70,2.25);
    \coordinate (labPicPos)  at (2.25,7.5);
    \coordinate (labWinPos) at (7.75,3.15);

    \draw[callout,shorten <=0.6pt,shorten >=0.8pt] (labCabinetPos)  -- (objCabinet.center);
    \draw[callout,shorten <=0.6pt,shorten >=0.8pt] (labPicPos)  -- (colPic.center);
    \draw[callout,shorten <=0.6pt,shorten >=0.8pt] (labWinPos) -- (objWin.center);

    \node[objlabel] at (labCabinetPos)  {green cabinet\\(forward)};
    \node[objlabel] at (labPicPos)  {several\\pictures\\(back)};
    \node[objlabel] at (labWinPos) {window and\\brown cabinet\\(right)};

    \def\boxSq{0.34}

    \coordinate (vC) at (5,5);

    \coordinate (v1) at ($(vC)+(-0.35,-0.35)$); 
    \coordinate (v2) at (vC);                   
    \coordinate (v3) at ($(vC)+(-1,+0)$);   

    \path[viewone, fill=selView!12,text=selView!80!black,] ($(v1)+(-\boxSq,-\boxSq)$) rectangle ($(v1)+(\boxSq,\boxSq)$);
    \node[viewnum] at (v1) {1};
    \draw[-{Latex[length=2.05mm]},black!75,line width=0.75pt]
      ($(v1)+(0,-\boxSq)$) -- ($(v1)+(0,-0.95)$);
    \node[fill=selView!12,text=selView!80!black,rounded corners=0.6pt,inner sep=0.25pt,anchor=west,align=center] at (2.75,4) {\textbf{selected}};

    \path[viewone] ($(v2)+(-\boxSq,-\boxSq)$) rectangle ($(v2)+(\boxSq,\boxSq)$);
    \node[viewnum] at (v2) {2};
    \draw[-{Latex[length=2.05mm]},black!75,line width=0.75pt]
      ($(v2)+(+\boxSq,0)$) -- ($(v2)+(+0.95,0)$);

    \path[viewone] ($(v3)+(-\boxSq,-\boxSq)$) rectangle ($(v3)+(\boxSq,\boxSq)$);
    \node[viewnum] at (v3) {3};
    \draw[-{Latex[length=2.05mm]},black!75,line width=0.75pt]
      ($(v3)+(0,+\boxSq)$) -- ($(v3)+(0,+0.95)$);

    \coordinate (egoC) at (9.00,9.00);
    \fill[selView] (egoC) circle (0.05);
    \draw[-{Latex[length=1.4mm]},line width=0.55pt] (egoC) -- ++(0.36,0);
    \draw[-{Latex[length=1.4mm]},line width=0.55pt] (egoC) -- ++(-0.36,0);
    \draw[-{Latex[length=1.4mm]},line width=0.55pt] (egoC) -- ++(0,-0.36);
    \draw[-{Latex[length=1.4mm]},line width=0.55pt] (egoC) -- ++(0,0.36);
    \node[anchor=west]  at ($(egoC)+(0.25,0)$) {R};
    \node[anchor=east]  at ($(egoC)+(-0.25,0)$) {L};
    \node[anchor=south] at ($(egoC)+(0,-0.2)$) {F};
    \node[anchor=north] at ($(egoC)+(0,0.2)$) {B};

  \end{scope}
\end{tikzpicture}%
                    }
                \end{minipage}

                \begin{minipage}[b][\qualCapH][c]{\linewidth}
                    \centering
                    {\tiny (c) CogMap with ego-directions}
                \end{minipage}
            \end{minipage}%
        }

        \par\smallskip

        \begin{tcolorbox}[
            width=\linewidth,
            colback=black!2,
            colframe=black!55,
            boxrule=0.45pt,
            arc=0.9pt,
            before skip=1pt,
            after skip=5pt,
            left=1.1mm,right=1.1mm,top=0.5mm,bottom=0.5mm,
            fontupper=\fontsize{6}{7}\selectfont\sffamily
        ]
        \textbf{Question:} If I am standing at the same spot and facing the same direction as shown in image 2, then I turn 90 degrees to the left, what is to my right?\\
        \textbf{Choices:} A. Green cabinet \quad \textbf{B. Window and brown cabinet (GT)} \quad C. Several pictures
        \end{tcolorbox}

        \begin{tcolorbox}[
            width=\linewidth,
            colback=red!3,
            colframe=red!70!black,
            boxrule=0.45pt,
            arc=0.9pt,
            left=1.1mm,right=1.1mm,top=0.5mm,bottom=0.5mm,
            fontupper=\fontsize{6}{7}\selectfont\sffamily
        ]
        \textbf{Qwen2.5-VL-3B-Instruct:} \texttt{<think>} \\

        Let's analyze this step-by-step:\\

        1. Understanding the initial positions:\\
           - Image 1 shows a room with a white ceiling, a white wall, and a door.\\
           - Image 2 shows the same room but rotated 90 degrees clockwise from Image 1.\\
           - Image 3 shows the room again, but rotated 90 degrees clockwise from Image 2.\\
        
        2. Identifying the objects in Image 2:\\
           - There is a green cabinet on the left side.\\
           - There is a window on the right side.\\
           - There is a brown cabinet below the window.\\
           - There are several pictures hanging on the wall above the brown cabinet.\\
        
        3. Determining the new orientation:\\
           - If you stand at the same spot and face the same direction as shown in Image 2, and then turn 90 degrees to the left, you would be facing the opposite direction of Image 2.\\
           - This means you would be facing the window and the brown cabinet.\\
        
        4. Identifying the object to your right:\\
           - When you face the window and the brown cabinet, the object to your right would be the green cabinet.\\
        
        Therefore, if you turn 90 degrees to the left from the position shown in Image 2, the object to your right would be the green cabinet.\\
        \textbf{Answer}: \textbf{\textcolor{red!75!black}{A}} \\
        \texttt{</think>}\\

        \texttt{<answer>}\textbf{\textcolor{red!75!black}{A. Green cabinet}}\texttt{</answer>}
        \end{tcolorbox}

        \begin{tcolorbox}[
            width=\linewidth,
            colback=green!4,
            colframe=green!45!black,
            boxrule=0.45pt,
            arc=0.9pt,
            left=1.1mm,right=1.1mm,top=0.5mm,bottom=0.5mm,
            fontupper=\fontsize{6}{7}\selectfont\sffamily
        ]
        \textbf{Ours:} \texttt{<think>}\\ 
        
        \textbf{(1) Selecting a view:}\\
        The question references Image 2 at \texttt{[5,5] facing="up"}. After turn left 90, the effective pose becomes \texttt{[5,5] facing="left"} (\textbf{\underline{Image 1}}). We should consider the viewpoint from \textbf{\underline{Image 1}}.\\
        
        \textbf{(2) Egocentric layout:}\\
        The question asks about ``right''. In the EgoMap, ``Window and brown cabinet'' has \texttt{ego\_direction="right"}, so it matches ``right''. \\
        
        \textbf{(3) Confirming in \textbf{\underline{Image 2:}}}\\
        Window and brown cabinet is visible, consistent with right in the EgoMap.\\
        
        \textbf{(4)} Therefore, the answer is \textbf{\textcolor{green!25!black}{``B. Window and brown cabinet''}}. \\
        \texttt{</think>}\\

        \texttt{<answer>}\textbf{\textcolor{green!25!black}{B. Window and brown cabinet}}\texttt{</answer>}
        \end{tcolorbox}

    \end{tcolorbox}
    \caption{
        \textbf{Qualitative comparison on MindCube-Tiny (Rotation).}
    }
    \label{fig:quali_mindcube_rotation_01}
\end{figure}

\begin{figure}[!ht]
    \centering

    \setlength{\qualTopA}{0.42\linewidth}
    \setlength{\qualTopB}{0.23\linewidth}
    \setlength{\qualTopC}{0.33\linewidth}
    \setlength{\qualTopH}{0.29\textwidth}  
    \setlength{\qualTopH}{0.25\textwidth}  
    \setlength{\qualCapH}{1.8em}
    \setlength{\qualImgH}{\dimexpr\qualTopH-\qualCapH\relax}
    \setlist[itemize]{label=\texttt{-}, leftmargin=1.6em, itemsep=0.15em, topsep=0pt}

    \begin{tcolorbox}[
        width=\linewidth,
        colback=black!1,
        colframe=black!70,
        boxrule=0.6pt,
        arc=1.2pt,
        before skip=0pt,
        after skip=3pt,
        left=1.5mm,right=1.5mm,top=1.1mm,bottom=1.1mm,
        title=\textcolor{white}{\texttt{\textbf{\scriptsize val\_Multi-view\_Reasoning\_67 (BLINK (MV))}}},
        fonttitle=\sffamily\bfseries,
        coltitle=black
    ]

        \makebox[\linewidth][c]{%
            \begin{minipage}[t][\qualTopH][t]{\qualTopA}
            \centering
        
            \begin{minipage}[c][\qualImgH][c]{\linewidth}
                \centering
                \setlength{\tabcolsep}{0pt}
                \renewcommand{\arraystretch}{0}
                \adjustbox{max width=0.98\linewidth, max totalheight=\qualImgH, keepaspectratio}{%
                    \begin{tabular}{@{}cc@{}}
                        \mvimg{}{1} &
                        \mvimg{}{2} \\  
                    \end{tabular}
                }
            \end{minipage}
        
            \begin{minipage}[b][\qualCapH][c]{\linewidth}
                \centering
                {\tiny (a) Multi-view images}
            \end{minipage}
        \end{minipage}\hfill

            \begin{minipage}[t][\qualTopH][t]{\qualTopB}
                \centering
                \begin{minipage}[c][\qualImgH][c]{\linewidth}
                    \centering
                    \adjustbox{max width=0.98\linewidth, max totalheight=\qualImgH, keepaspectratio}{%
                        \includegraphics[trim={6 6 6 6},clip]{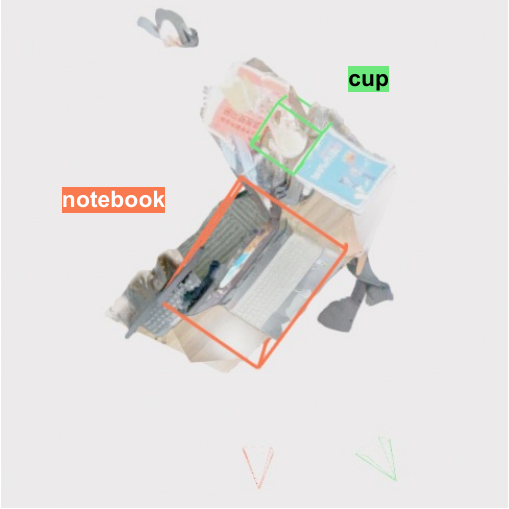}%
                    }
                \end{minipage}

                \begin{minipage}[b][\qualCapH][c]{\linewidth}
                    \centering
                    {\tiny (b) Scene from VFM}
                \end{minipage}
            \end{minipage}\hfill
            \begin{minipage}[t][\qualTopH][t]{\qualTopC}
                \centering
                \begin{minipage}[c][\qualImgH][c]{\linewidth}
                    \centering
                    \adjustbox{max width=0.97\linewidth, max totalheight=\qualImgH, keepaspectratio}{%
\begin{tikzpicture}[font=\fontsize{5.4}{6.0}\selectfont\sffamily]
  \begin{scope}[x=0.46cm,y=-0.46cm]
    \foreach \x in {0,...,10} {\draw[gray!40,line width=0.28pt] (\x,0) -- (\x,10);}
    \foreach \y in {0,...,10} {\draw[gray!40,line width=0.28pt] (0,\y) -- (10,\y);}
    \draw[black,line width=0.75pt] (0,0) rectangle (10,10);

    \definecolor{selView}{RGB}{226,140,28} 

    \definecolor{colNotebook}{RGB}{255, 92, 135}      
    \definecolor{colCup}{RGB}{50, 180, 90}      

    \tikzset{
      objdot/.style={draw=black!60,line width=0.35pt,circle,minimum size=4.4pt,inner sep=0pt},
      objlabel/.style={draw=black!45,fill=white,rounded corners=1.0pt,inner sep=1.0pt,align=center},
      callout/.style={black!55,line width=0.34pt,dash pattern=on 1pt off 1pt},
      viewone/.style={draw=black, fill=white, line width=0.45pt},
      viewselect/.style={draw=black, fill=selView!12, line width=0.45pt},
      viewnum/.style={font=\bfseries\fontsize{5.0}{5.0}\selectfont\sffamily}
    }

    \node[objdot,fill=colNotebook]    (objNotebook)    at (5,5) {};
    \node[objdot,fill=colCup] (objCup) at (5,3) {};

    \coordinate (labNotebookPos)    at (3.25,4.45);
    \coordinate (labCupPos)   at (7.55,3.85);

    \draw[callout,shorten <=0.6pt,shorten >=0.8pt] (labNotebookPos)    -- (objNotebook.center);
    \draw[callout,shorten <=0.6pt,shorten >=0.8pt] (labCupPos)   -- (objCup.center);

    \node[objlabel] at (labNotebookPos)    {notebook\\(forward)};
    \node[objlabel] at (labCupPos)   {cup\\(forward)};

    \def\boxSq{0.34}

    \path[viewselect] (5-\boxSq,7-\boxSq) rectangle (5+\boxSq,7+\boxSq);
    \node[viewnum] at (5,7) {1};
    \node[fill=selView!12,text=selView!80!black,rounded corners=0.6pt,inner sep=0.25pt,anchor=west]
      at (4.25,8.00) {\textbf{selected}};
    \draw[-{Latex[length=2.25mm]},black!75,line width=0.95pt]
      (5,7-\boxSq) -- (5,5.75);

    \path[viewone] (7-\boxSq,7-\boxSq) rectangle (7+\boxSq,7+\boxSq);
    \node[viewnum] at (7,7) {2};
    \draw[-{Latex[length=2.05mm]},black!75,line width=0.75pt]
      (7-\boxSq,7-\boxSq) -- (6.25,6.25);

    \coordinate (egoC) at (9.00,9.00);
    \fill[selView] (egoC) circle (0.05);
    \draw[-{Latex[length=1.4mm]},line width=0.55pt] (egoC) -- ++(0.36,0);
    \draw[-{Latex[length=1.4mm]},line width=0.55pt] (egoC) -- ++(-0.36,0);
    \draw[-{Latex[length=1.4mm]},line width=0.55pt] (egoC) -- ++(0,-0.36);
    \draw[-{Latex[length=1.4mm]},line width=0.55pt] (egoC) -- ++(0,0.36);
    \node[anchor=west]  at ($(egoC)+(0.25,0)$) {R};
    \node[anchor=east]  at ($(egoC)+(-0.25,0)$) {L};
    \node[anchor=south] at ($(egoC)+(0,-0.2)$) {F};
    \node[anchor=north] at ($(egoC)+(0,0.2)$) {B};

  \end{scope}
\end{tikzpicture}%
                    }
                \end{minipage}

                \begin{minipage}[b][\qualCapH][c]{\linewidth}
                    \centering
                    {\tiny (c) CogMap with ego-directions}
                \end{minipage}
            \end{minipage}%
        }

        \par\smallskip

        \begin{tcolorbox}[
            width=\linewidth,
            colback=black!2,
            colframe=black!55,
            boxrule=0.45pt,
            arc=0.9pt,
            before skip=1pt,
            after skip=5pt,
            left=1.1mm,right=1.1mm,top=0.5mm,bottom=0.5mm,
            fontupper=\fontsize{6}{7}\selectfont\sffamily
        ]
        \textbf{Question:} The images are frames from a video. The first image is from the beginning of the video, and the second image is from the end. Is the camera moving left or right when shooting the video? Select between A and B. \\
        \textbf{Choices:} A. left \quad \textbf{B. right (GT)}
        \end{tcolorbox}

        \begin{tcolorbox}[
            width=\linewidth,
            colback=red!3,
            colframe=red!70!black,
            boxrule=0.45pt,
            arc=0.9pt,
            left=1.1mm,right=1.1mm,top=0.5mm,bottom=0.5mm,
            fontupper=\fontsize{6}{7}\selectfont\sffamily
        ]
        \textbf{Qwen2.5-VL-3B-Instruct:} \texttt{<think>}\\
        Let's break down this analysis:\\
        
          \textbf{Image Description:}
          \begin{itemize}
            \item First Image:
            \begin{itemize}
              \item There is an open laptop with its screen facing forward.
              \item To the left of the laptop, there appears to be another monitor displaying some content.
            \end{itemize}
        
            Second Image:
            \begin{itemize}
              \item Similar setup but now we see more details such as someone sitting at the desk wearing jeans and sneakers.
              \item Additional items visible include water bottles placed near the laptops.\\
            \end{itemize}
          \end{itemize}
          \textbf{Analysis:}
          \begin{itemize}
            \item To determine if the camera moves left or right between these two images, let’s consider several factors including perspective changes and relative positions observed across both shots.\\
          \end{itemize}
          \textbf{Perspective Changes:}
          \begin{itemize}
            \item In the initial frame, only one person seems present based solely on their partial view through the laptop screens.
            \item By the time it transitions into the next shot, additional elements come into focus. Notably showing part of another individual seated closer to us than before due likely to increased depth perception caused by movement towards them.\\
          \end{itemize}
          \textbf{Relative Position Shifts:}
          \begin{itemize}
            \item Initially focused mainly on objects behind individuals working; gradually shifts attention toward foreground figures, indicating a horizontal panning motion. The consistent framing implies a lateral shift along the X-axis, ruling out vertical rotation. The gradual repositioning from background to foreground suggests the camera is panning to the left.\\
          \end{itemize}
         \textbf{Answer:} \textbf{\textcolor{red!75!black}{A}}
        
        \texttt{</think>}\\

        \texttt{<answer>}\textbf{\textcolor{red!75!black}{A. left}}\texttt{</answer>}
        \end{tcolorbox}

        \begin{tcolorbox}[
            width=\linewidth,
            colback=green!4,
            colframe=green!45!black,
            boxrule=0.45pt,
            arc=0.9pt,
            left=1.1mm,right=1.1mm,top=0.5mm,bottom=0.5mm,
            fontupper=\fontsize{6}{7}\selectfont\sffamily
        ]
        \textbf{Ours:} \texttt{<think>}\\ 
        
        \textbf{(1) Selecting a view:}\\
        The question asks the movement direction from Image 1 to Image 2, so we should consider the viewpoint at \textbf{\underline{Image 1.}}\\
        
        \textbf{(2) Egocentric layout:}\\
        In the EgoMap, ``notebook'' has \texttt{ego\_direction}=\texttt{``forward''}. Also, the viewpoint at Image 2 is at the position \texttt{(7,7)}. This corresponds to the ``right''.\\
        
        \textbf{(3) Confirming in \textbf{\underline{Image 2:}}}\\
        ``Notebook'' appears on the left side of the cup, whereas it is on the ``right'' side in Image 1. This is consistent with a rightward camera movement.\\
        
        \textbf{(4)} Therefore, the answer is \textbf{\textcolor{green!25!black}{``B. right''}}. \\
        \texttt{</think>}\\

        \texttt{<answer>}\textbf{\textcolor{green!25!black}{B. right}}\texttt{</answer>}
        \end{tcolorbox}

    \end{tcolorbox}
    \caption{
        \textbf{Qualitative comparison on BLINK (MV).}
        The frozen Qwen2.5-VL-3B-Instruct fails to choose correct answer despite lengthy reasoning.
        In contrast, our method explicitly selects the appropriate reference view and resolves the camera motion with an ego-consistent reasoning trace, achieving the correct answer.
    }
    \label{fig:quali_blink_01}
\end{figure}

\subsection{Additional Qualitative Results}
We provide additional qualitative examples on all three benchmarks to illustrate how the proposed framework constructs intermediate spatial representations and performs grounded reasoning under diverse visual configurations.
For MindCube, we show representative cases from both the \textit{Among} and \textit{Rotation} settings in~\Cref{fig:quali_mindcube_among_01,fig:quali_mindcube_rotation_01}, highlighting how the model combines allocentric scene understanding with viewpoint-dependent egocentric grounding.
Finally,~\Cref{fig:quali_blink_01} shows qualitative results on BLINK (MV), demonstrating that the proposed structured reasoning format also generalizes to multi-view perception problems beyond the original training benchmark.
Overall, these examples qualitatively support that our method produces interpretable intermediate reasoning steps while remaining robust across different benchmark settings.

\subsection{Training Plot and Reward Plot Analysis}

\begin{figure}[!ht]
    \centering
    \includegraphics[width=0.95\linewidth]{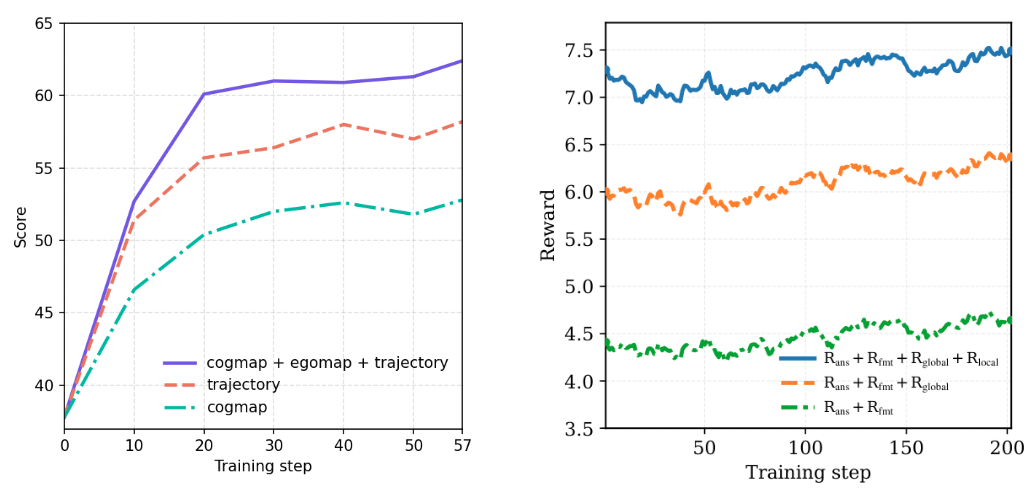}
    \caption{
    \textbf{Training dynamics under different SFT and GRPO settings.}
    \textbf{Left:} During SFT, the average score increases and then saturates across settings,  while the full setting reaches the highest plateau compared to variants only with trajectory supervision or spatial maps.
    \textbf{Right:} GRPO optimization proceeds stably under different reward settings, as all curves increase and then stabilize.
    }
  \label{fig:reward_curve}
\end{figure}

We analyze training dynamics for both SFT and GRPO in our method.
For SFT, we track the validation accuracy on MindCube-Tiny for up to 57 training steps.
The left plot in~\Cref{fig:reward_curve} compares three supervision settings: allocentric cognitive map (CogMap) only, trajectory only, and the full setting with both spatial maps and trajectory supervision.
CogMap-only supervision converges to an average score of about 53, while trajectory-only supervision reaches about 58.
The full setting achieves the highest plateau at about 62, which shows that spatial-map supervision and trajectory supervision are complementary.
In particular, adding egocentric guidance with the trajectory of viewpoint selection leads to the greatest improvement in the final score.

For GRPO, the right plot in~\Cref{fig:reward_curve} shows the mean reward over 200 training steps.
We apply EMA smoothing to the step-wise reward because the raw reward is noisy across steps.
We compare three reward settings: (i) the full reward, (ii) without $\mathcal{R}_{\text{local}}$, and (iii) without both $\mathcal{R}_{\text{global}}$ and $\mathcal{R}_{\text{local}}$.
Across all settings, the reward increases and then stabilizes, which indicates that GRPO optimization remains stable under each reward definition.
The three curves follow a similar convergence pattern, but the full reward consistently stays above the ablated variants throughout training.
This trend supports our reward design, as combining global and local rewards leads to the highest reward level under the same GRPO procedure.

\subsection{Effects of Egocentric Grounding}
\begin{table}[t]
\centering
\scriptsize
\caption{\textbf{Analysis on effects of egocentric representation in inference using frozen MLLM (Qwen2.5-VL-3B-Instruct).}}
\label{tab:mindcube_frozen_ego}
\resizebox{0.95\textwidth}{!}{
\setlength{\tabcolsep}{3.6pt}
\renewcommand{\arraystretch}{1.06}

\begin{threeparttable}

\begin{tabular}{l cc ccc cccc}
\toprule
\multirow{2}{*}{\textbf{Config.}} &
\multicolumn{2}{c}{\textbf{Input}} &
\multicolumn{3}{c}{\textbf{Output}} &
\multicolumn{4}{c}{\textbf{MindCube-Tiny}} \\
\cmidrule(l){2-3}\cmidrule(l){4-6}\cmidrule(l){7-10}
& \textbf{CogMap} & \textbf{EgoMap}
& \textbf{CogMap} & \textbf{EgoMap} & \textbf{Reasoning}
& \textbf{Rotation} & \textbf{Among} & \textbf{Around} & \textbf{Overall} \\
\midrule

\texttt{CogMap-FFR-Out}
& \xmark & \xmark & \cmark & \xmark & \cmark
& 25.00 & 39.67 & 58.40 & 41.33 \\

\texttt{CogMap-In-FFR-Out}
& \cmark & \xmark & \xmark & \xmark & \cmark
& 37.00 & 41.67 & 44.40 & 41.43 \\

\texttt{EgoMap-FFR-Out}
& \xmark & \xmark & \xmark & \cmark & \cmark
& 30.00 & 39.33 & 54.80 & 41.24 \\

\texttt{EgoMap-In-FFR-Out}
& \xmark & \cmark & \xmark & \xmark & \cmark
& 41.00 & 43.17 & 54.80 & 45.52 \\

\texttt{CogMap-EgoMap-FFR-Out}
& \xmark & \xmark & \cmark & \cmark & \cmark
& 33.50 & 45.50 & 54.80 & 45.43 \\

\texttt{CogMap-In-EgoMap-FFR-Out}
& \cmark & \xmark & \xmark & \cmark & \cmark
& 35.00 & 43.67 & 52.80 & 44.19 \\

\texttt{CogMap-EgoMap-In-FFR-Out}
& \cmark & \cmark & \xmark & \xmark & \cmark
& 55.50 & 51.17 & 52.00 & \textbf{52.19} \\

\bottomrule
\end{tabular}
\end{threeparttable}
}
\end{table}

We investigate the effect of egocentric grounding in an inference-only setting using a frozen MLLM (Qwen2.5-VL-3B-Instruct), without SFT or RL.
We vary whether the model is (i) provided with an allocentric cognitive map (\texttt{CogMap}) or an egocentric map (\texttt{EgoMap}) as \textbf{inputs} (\texttt{-In}) and (ii) instructed to \textbf{generate} the maps as part of its \textbf{outputs} (\texttt{-Out}) with free-form reasoning (\texttt{FFR}).

As the results on MindCube-Tiny are summarized in \Cref{tab:mindcube_frozen_ego}, using only an allocentric cognitive map does not meaningfully change performance depending on whether it is provided as input or generated as output.
Specifically, \texttt{CogMap-FFR-Out} and \texttt{CogMap-In-FFR-Out} achieve nearly identical overall scores with 41.33 and 41.43, respectively, and this supports that grounding an allocentric map alone is not the main driver of improvement in this zero-shot setting. In contrast, introducing an egocentric representation yields a clear gain.
Providing an egocentric map as input (\texttt{EgoMap-In-FFR-Out}) improves the overall score to 45.52, and jointly providing both \texttt{CogMap} and \texttt{EgoMap} as inputs (\texttt{CogMap-EgoMap-In-FFR-Out}) further increases the overall score to 52.19.
These results suggest that egocentric grounding is important for spatial reasoning under viewpoint-dependent queries, and motivate our method to explicitly incorporate egocentric representations as grounding signals during training.

\subsection{Failure Cases}
Despite the gains from map-grounded supervision and dense rewards, our method still has a clear limitation on scenes that violate the static-world assumption underlying our cognitive-map construction.
In particular, the current framework assumes that the multi-view or multi-frame inputs can be integrated into a single coherent scene layout with stable object positions, so that camera motion can be inferred relative to a largely static environment.
This assumption breaks when salient foreground objects move independently of the camera.
In such cases, the model may incorrectly use a moving object as a spatial reference and attribute the apparent motion to ego-motion of the camera, leading to a wrong answer.
For example, when a white car becomes closer across frames~\cite{yang2025mmsi}, the model predicts that the camera is moving forward, although the correct interpretation is that the object itself is moving and the camera is moving backward.
This failure suggests that our current map-based reasoning is effective primarily for quasi-static scenes, but is less reliable when dynamic objects introduce ambiguity between object motion and camera motion.
Addressing this limitation would require explicit motion disentanglement or dynamic scene representations beyond the static allocentric-egocentric pipeline considered in this work.

\end{document}